\pdfoutput=1
\documentclass[10pt,twocolumn]{article}
\usepackage{cvpr}
\usepackage{times}
\usepackage{caption}
\usepackage{epsfig}
\usepackage{graphicx}
\usepackage{amsmath}
\usepackage{amssymb}
\usepackage{makecell}
\usepackage{algorithm}
\usepackage[noend]{algpseudocode}
\usepackage{wrapfig,booktabs}
\usepackage{color}
\usepackage[normalem]{ulem}
\usepackage[table]{xcolor}
\usepackage{pifont}
\usepackage{multirow}
\usepackage{mathtools}
\usepackage{dsfont}
\usepackage{comment}
\usepackage{tabularx}

\usepackage[pagebackref=true,breaklinks=true,colorlinks,bookmarks=false]{hyperref}
\cvprfinalcopy 

\def\cvprPaperID{4} 

\ifcvprfinal\fi

\begin{document}

\title{Finding Berries: Segmentation and Counting of Cranberries using Point Supervision and Shape Priors}


\author{\large Peri Akiva\textsuperscript{1} \hspace{0.4cm} Kristin Dana\textsuperscript{1} \hspace{0.4cm} Peter Oudemans\textsuperscript{2} \hspace{0.4cm} Michael Mars\textsuperscript{2}\\
\textsuperscript{1}Department of Computer and Electrical Engineering, Rutgers University\\
\textsuperscript{2}Department of Plant Biology, Rutgers University\\
{\tt\small \{peri.akiva, kristin.dana\}@rutgers.edu \hspace{0.4cm} \{oudemans, mm2784\}@njaes.rutgers.edu}
}

\maketitle

\begin{abstract}
Precision agriculture has become a key factor for increasing crop yields by providing essential information to decision makers. In this work, we present a deep learning method for simultaneous segmentation and counting of cranberries to aid in yield estimation and sun exposure predictions. Notably, supervision is done using low cost center point annotations. The approach, named Triple-S Network,  incorporates a three-part loss  with shape priors to promote better fitting to  objects of known shape typical in agricultural scenes.  Our results improve overall segmentation performance by more than 6.74\% and counting results by 22.91\% when compared to state-of-the-art. To train and evaluate the network, we have collected the CRanberry Aerial Imagery Dataset (CRAID), the largest dataset of aerial drone imagery from cranberry fields. This dataset will be made publicly available. 
\end{abstract}
\vspace{-0.40cm}
\vspace{-0.2cm}
\thispagestyle{empty}
\section{Introduction}
\begin{figure*}
\centering
\setlength\tabcolsep{1.5pt}
\def\arraystretch{1}
\begin{tabularx}{\textwidth}{cc}
\centering
    \includegraphics[width=0.489\linewidth]{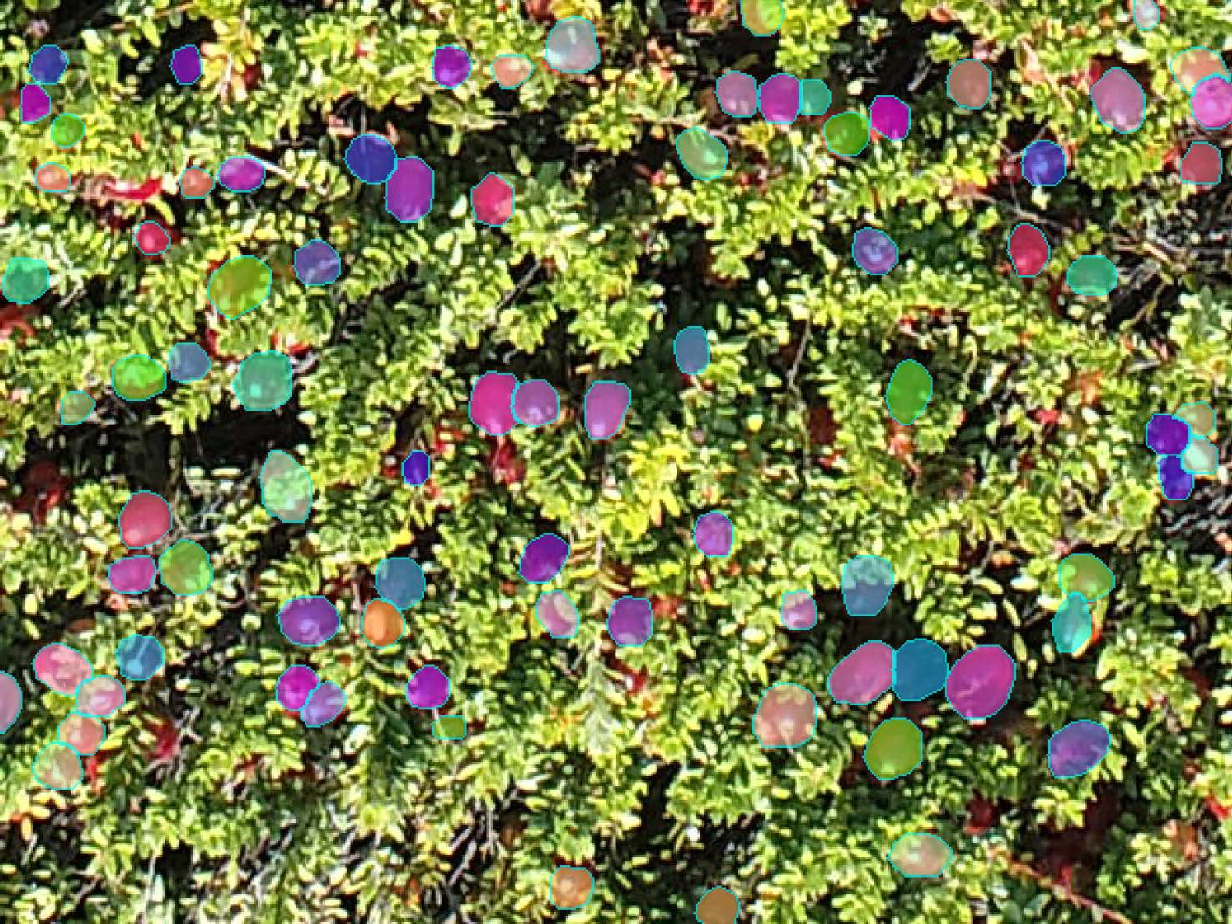} &  \includegraphics[width=0.489\linewidth]{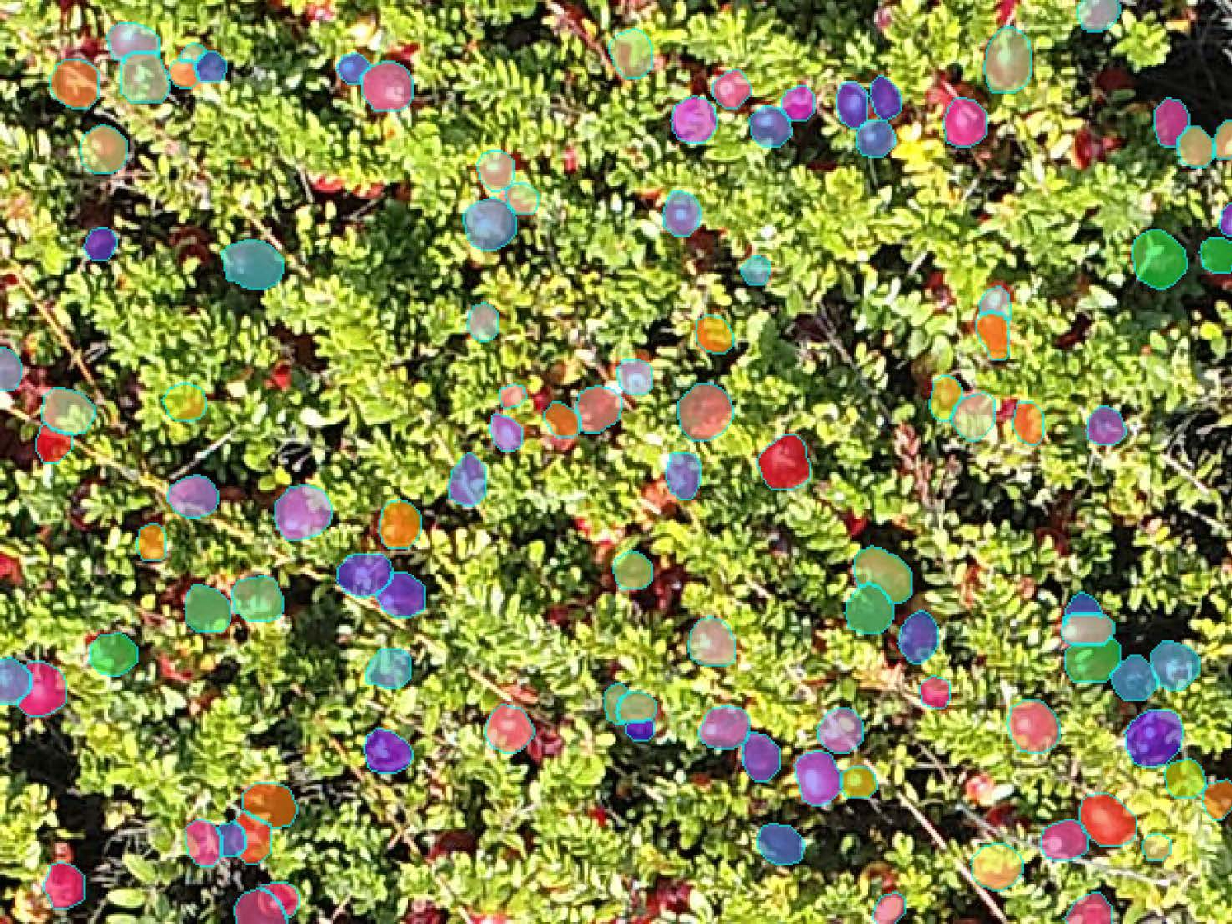}\\
    \includegraphics[width=0.489\linewidth]{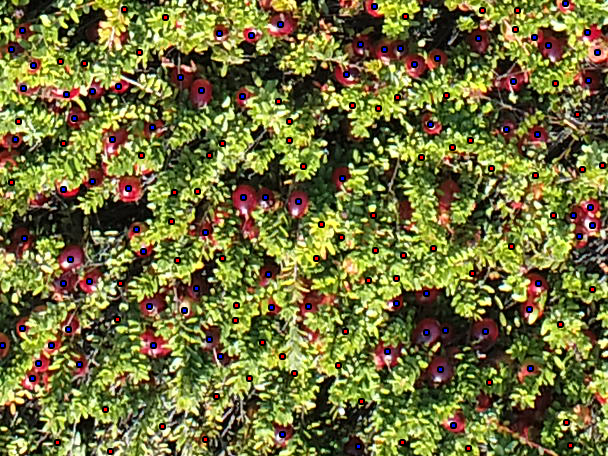} &  \includegraphics[width=0.489\linewidth]{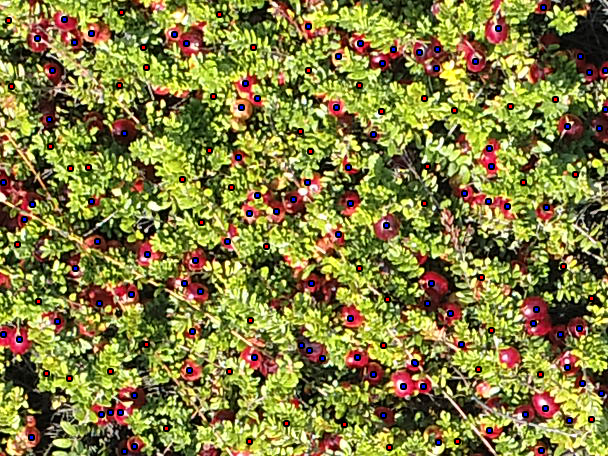}\\
\end{tabularx}
\vspace{-0.26cm}
\caption{Examples of CRAID images with overlaid berry-wise and point-wise ground truth masks. Red and blue dots represent cranberry and background examples. Colors in ground truth masks are random and are used to represent instances. Colors may repeat. Best viewed in color and zoomed.}
\vspace{-0.56cm}
\label{dataset}
\end{figure*}

The challenges of agriculture presents new opportunities for computer vision methods. Evaluation of crop health, sun exposure and anticipated yields using computational algorithms leads to new methods of farming and resource management. Automated segmentation and counting provides a method of determining value of produce and anticipated profits, as well as optimizing irrigation and water management. 
Current yield estimation methods rely on data from previous years or manual measurements of small regions. These processes limit the accuracy  of predicted yield since weather can be vastly different in consequent years, and randomly sampled measurements may be skewed and costly. Recent studies \cite{roper2006physiology,kerry2017investigating,pelletier2016reducing} show that lack of informed decision-making 
is a significant cause of lost produce. 
For example, \cite{pelletier2016reducing} investigates the effect of sudden changes in air temperature from heat waves causing regions with up to 100\% yield loss 
due to a combination of heat stress and water stress. 
Managing water resources requires balancing the tradeoff of irrigation costs and yield risk. 
Our goal is to build a non-invasive vision-based crop analysis platform that segments and counts exposed berries, and can serve as a low-cost automated tool for estimating yield and sun exposure.
\begin{figure}[t!]
\centering
\includegraphics[width=\linewidth]{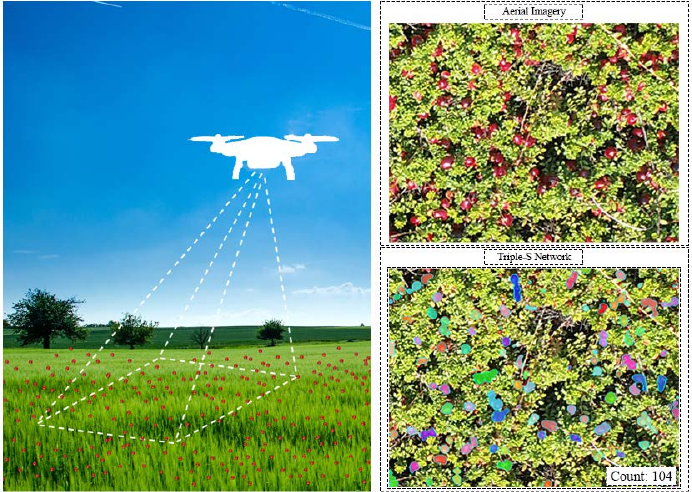}
\vspace{-0.55cm}
\caption{Overview of pipeline. Left: scene illustration of data collection stage. Top right: image captured by the drone. Bottom right: segmentation and count outputs of our Triple-S network. Colors in prediction mask are random and are used to represent instances (colors may repeat). Best viewed in color and zoomed.}
\vspace{-0.65cm}
\label{teaserfigure}
\end{figure}
Current precision agriculture state-of-the-art (SOTA) methods utilize ground vehicles, high resolution cameras, lidars, multispectral sensors, and thermal sensors to automate this process, creating  more accurate and cost effective solutions to yield and sun exposure estimations. These instruments are utilized in recent precision agriculture work seeking to detect and segment fruits and weeds \cite{song_glasbey_horgan_polder_dieleman_heijden_2014, bargoti_underwood_2017,bargoti_underwood_2017_2, kestur_meduri_narasipura_2019,lottes2018fully}. However, these sensors are expensive, require specialized knowledge to operate, and often need close proximity to the targeted objects. Systems such as in \cite{bargoti_underwood_2017,bargoti_underwood_2017_2} also require the use of invasive ground vehicle and trained drivers which further increase the cost of the system.
Surveys of remote sensing
with unmanned aerial vehicles (UAVs)
\cite{maes2019perspectives} highlight the importance of non-invasive systems in precision agriculture.
Detection of fruit stress and pathogens with UAVs  \cite{garcia2013comparison,albetis2017detection,calderon2013high,lopez2016early,tetila2017identification,hunt2017detection,csillik2018identification} typically require expensive hyperspectral and thermal sensors. Additionally, most UAV methods focus on fruit crop images which are simpler compared to cranberry crops that have  numerous occluding leaves in the canopy imagery (see Figure~\ref{dataset}).
Our approach seeks to count and segment cranberries in RGB images
collected by non-invasive equipment.
Recent segmentation methods require training algorithms using pixel-wise ground truth obtained manually \cite{zhou2017scene, lin2014microsoft, Everingham10}, but such ground truth is expensive to obtain. We develop a novel method using only {\it point-wise annotations} which are an order magnitude cheaper than full pixel-wise supervision \cite{bearman2016s}.
Our approach pairs the point-click annotations with additional shape and convexity cues to produce instance segmentation results.  



The primary contributions of this work are as follows:
\begin{itemize}
    \item We propose a method named Triple-S Network that encourages shape-specific instance segmentation predictions for small, many-object scenes driven by known shape priors and point supervision.
    \item We present a a selective watershed algorithm that uses both negative and positive seeds for selective segmentation masks generation.
    \item We outperform SOTA point supervision semantic segmentation and counting methods on our dataset.
    \item We provide the largest publicly available dataset of aerial images of cranberry crops with pixel-wise and center point annotations named Cranberry Aerial Imagery Dataset (CRAID).
\end{itemize}

\section{Related Work}

\paragraph{Computer Vision in Agriculture.} 
Early precision agriculture using aerial imagery began in  the 1980's and includes Soil Teq's field soil fertility mapping system for crops \cite{zheng_schreier_1988} using spectral features.
Studies that correlate precision agriculture to higher yields of crops \cite{cassman_1999,stafford_2000} motivated researchers to use computer vision for new ways to measure, survey, and estimate yield of crops. Early work in this domain utilizes colors, shapes \cite{chaisattapagon_1995,perez_lopez_benlloch_christensen_2000,el-faki_zhang_peterson_2000,franz_gebhardt_unklesbay_1991}, reflection levels \cite{vrindts_baerdemaeker_1999,hunt2017detection}, and multi-spectral features \cite{feyaerts_gool_2001,di2016unmanned,garcia2013comparison} to detect and evaluate fruits, wheat, and weeds. These methods apply image pre-processing techniques such as contrast and thresholding with machine learning algorithms such as k-nearest neighbors, decision trees, and support vector machines. While  early models may perform well under controlled conditions and small datasets, they fail to generalize over diverse and noisy inputs common in real world applications. 
More recent models incorporate deep learning algorithms to generalize over different environments. Song et al.\ \cite{song_glasbey_horgan_polder_dieleman_heijden_2014} propose patch-wise fruit classification, using a combination of color classifiers for key-point extraction and fixed patches around each key-point. These patches are then classified as either fruit or non-fruit images. Bargoti and Underwood \cite{bargoti_underwood_2017} propose a segmentation model using fully convolutional networks (FCN) \cite{long_shelhamer_darrell_2015} using fully supervised RGB images and meta-data pertaining to camera angle, camera location, type of tree captured, and weather conditions. The model's output is then processed by the watershed algorithm \cite{meyer_1994} to produce separable regions used for fruit counting. Combining elements from \cite{bargoti_underwood_2017,song_glasbey_horgan_polder_dieleman_heijden_2014}, Kestur et al.\ \cite{kestur_meduri_narasipura_2019} use $200 \times 200$ input patches to a modified fully convolutional network to generate masks patch-wise and stitch them together. 
A similar FCN model \cite{lottes2018fully} is trained on near infra red (NIR) and RGB sequences.
These models, however, offer limited performance with high operation costs, requiring ground vehicles and fully supervised ground truth data. 
In addition to significant prior work using convolutional networks in fruit imaging, automated weed detection methods use similar deep learning models \cite{yu2019deep} to distinguish between weed types in RGB images taken from a fixed altitude.
The model is trained with image level labels that indicate weed type and is tested on input's 9 sub-images, providing patch-wise weed predictions. 
%

%
\vspace{-0.5cm}
\paragraph{Weakly Supervised Segmentation and Counting.} Instance segmentation seeks to not only find the class of each pixel, but also the object instance it belongs to, which indirectly provides object counts in a given scene. Initial development in this task domain is derived from R-CNN \cite{girshick2014rich, girshick2015fast}, utilizing proposal based segmentation.
More recent work attempts to minimize the amount of supervision while producing similar performance. The work of \cite{dai2015boxsup,khoreva2017simple} first suggested semantic segmentation methods using bounding boxes, followed by \cite{song2019box} to showcase SOTA segmentation using predefined class-wise filling rates. While these methods perform well on everyday scenes, they require bounding box annotations which are more expensive than point annotations, and are computationally demanding, utilizing region proposal networks \cite{girshick2014rich,ren2015faster} to generate proposal masks for sets of anchors originating at each pixel. Less supervised methods  \cite{zhou2018weakly, laradji2019masks} aim to segment scenes based on image level labels. PRM (Peak Response Map) \cite{zhou2018weakly} makes use of class peak response to obtain instance aware visual cues from given inputs. The network generates peak response maps by backpropagating local peaks found in intermediate attention maps. While \cite{zhou2018weakly} reports SOTA performance on common datasets, the method exhibits increasing errors when the size and number of objects in the scene increase, which is confirmed by our experiments. Additionally, the network requires pre-processed segment proposals generated by a separate region proposal network.  Expanding on PRM \cite{zhou2018weakly}, \cite{laradji2019masks} refines its output to create pseudo masks used as ground truth to a fully supervised Mask R-CNN \cite{he2017mask} model that is robust to noisy masks. Similar to PRM, \cite{laradji2019masks} also requires a separate object proposal network to generate its pseudo masks while facing similar difficulties with small, many object scenes. While instance segmentation finds count indirectly, \cite{ribera2019locating} chooses to directly find counts and locations using center point annotations. This method  introduces the Weighted Hausdorff Distance loss to encourage better localization, while regressing over the joint latent features and network output to directly estimate counts. The work closest to our approach is LC-FCN (Localization-based Counting FCN) \cite{laradji2018blobs}, in which the model aims to detect regions in objects using center point annotations driven by a loss function that encourages object boundaries, point localization, and overall image loss.
\begin{figure*}[t!]
\centering
\includegraphics[width=0.99\linewidth]{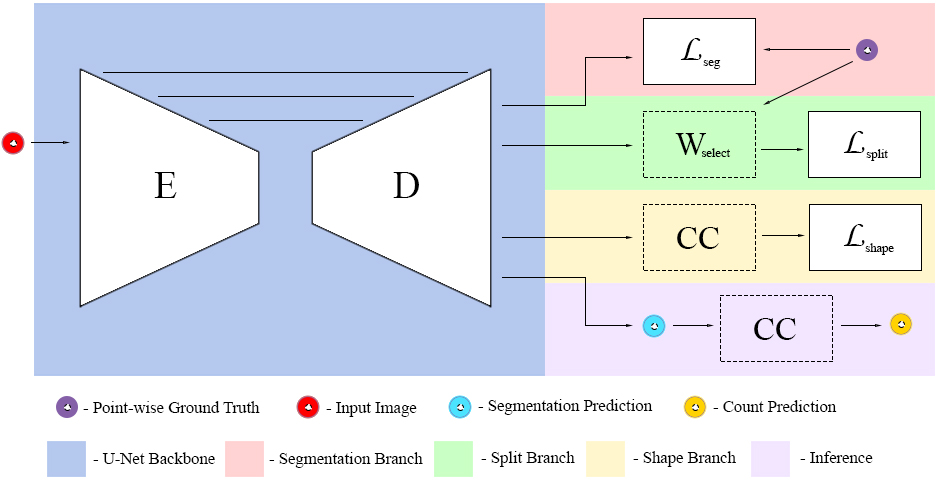}
\vspace{-0.33cm}
\caption{The network architecture of our proposed method. U-Net \cite{ronneberger2015u} with encoder E and decoder D is used as a backbone to generate masks guided by segmentation loss$\mathcal{L}_{Seg}$, split loss $\mathcal{L}_{Split}$ and shape loss $\mathcal{L}_{Shape}$. Our selective watershed algorithm, $\text{W}_{\text{select}}$, is used to better define expandable regions and boundaries in the predicted mask before computing the split loss. The shape loss branch first determines connected components, noted as CC, in the prediction mask before calculating individual shape loss. During inference, the predicted mask is obtained directly from the U-Net, and the count is calculated by the number of connected components present in the predicted segmentation.}
\vspace{-0.55cm}
\label{model}
\end{figure*}
The split loss used in LC-FCN utilizes the set of pixels representing the boundaries of objects obtained by the watershed algorithm and is calculated for individual blobs and overall image. In contrast, our split loss considers the set of pixels representing the possible regions objects can expand to without crossing to neighboring objects and is calculated against the prediction mask (see Figure~\ref{selectivewatershed} for visual comparison).
This approach aims to penalize the model if the predicted area is too small, while LC-FCN only penalizes the model if an object crosses a boundary. Additionally, we better constitute object borders using our selective watershed algorithm which uses negative and positive ground truth annotations to define positive and negative regions, unlike LC-FCN which only uses positive ground truth annotations in it watershed split loss.


\vspace{-0.2cm}
\section{CRAID: CRanberry Aerial Imagery Dataset}
\vspace{-0.1cm}
\subsection{Data Collection}
We collect 21,436 cranberry images of size $456 \times 608$ to create the largest repository of aerial RGB imagery of cranberry fields which we name CRAID. 
Images were collected using a Phantom 4 drone from a small range of altitudes with manually fixed camera settings: 100 ISO, 1/240 shutter speed, and 5.0 F-Stop. Data was acquired at weekly intervals to capture albedo variations in cranberries, starting at early bloom to post harvest. Drone trajectory is fixed throughout the collection season using initial randomly sampled path points at each cranberry bed. Before each recording session, a set of images of a checkerboard from different angels is captured for camera calibration purposes.
\begin{figure*}[t!]
\setlength\tabcolsep{1.5pt}
\def\arraystretch{1}
\centering
\begin{tabularx}{\textwidth}{ccc}
    \includegraphics[width=0.33\linewidth]{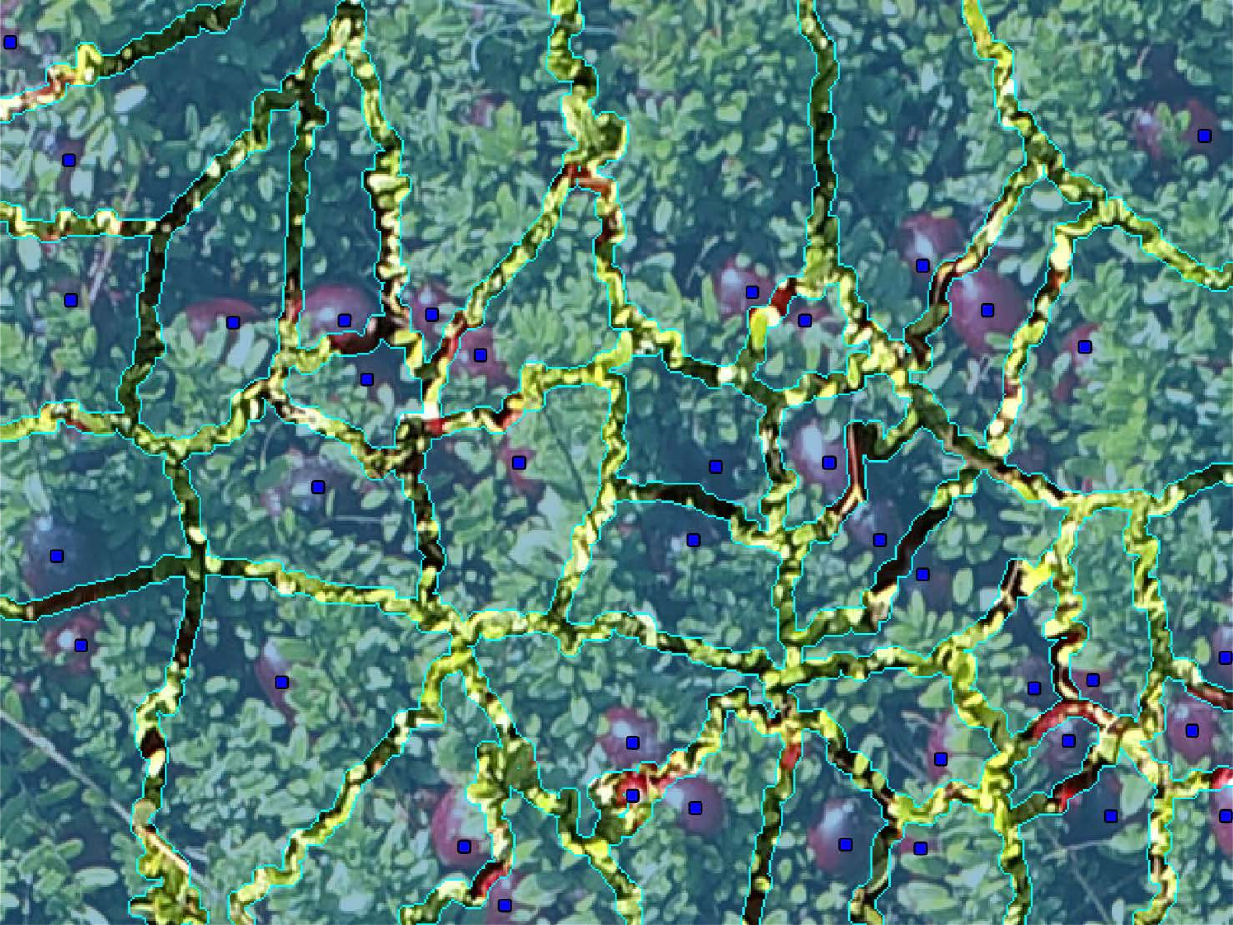} &   \includegraphics[width=0.33\linewidth]{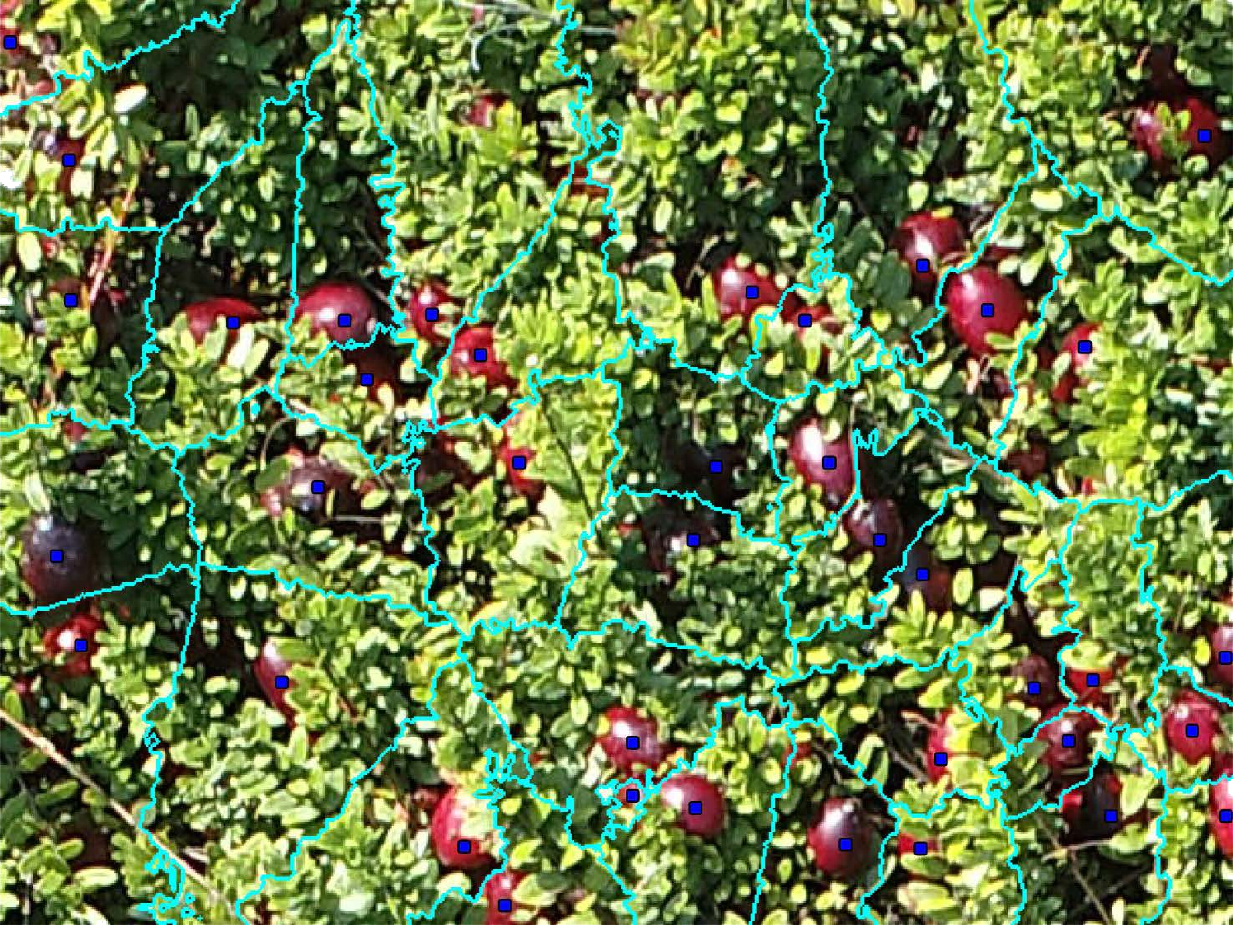} & 
    \includegraphics[width=0.33\linewidth]{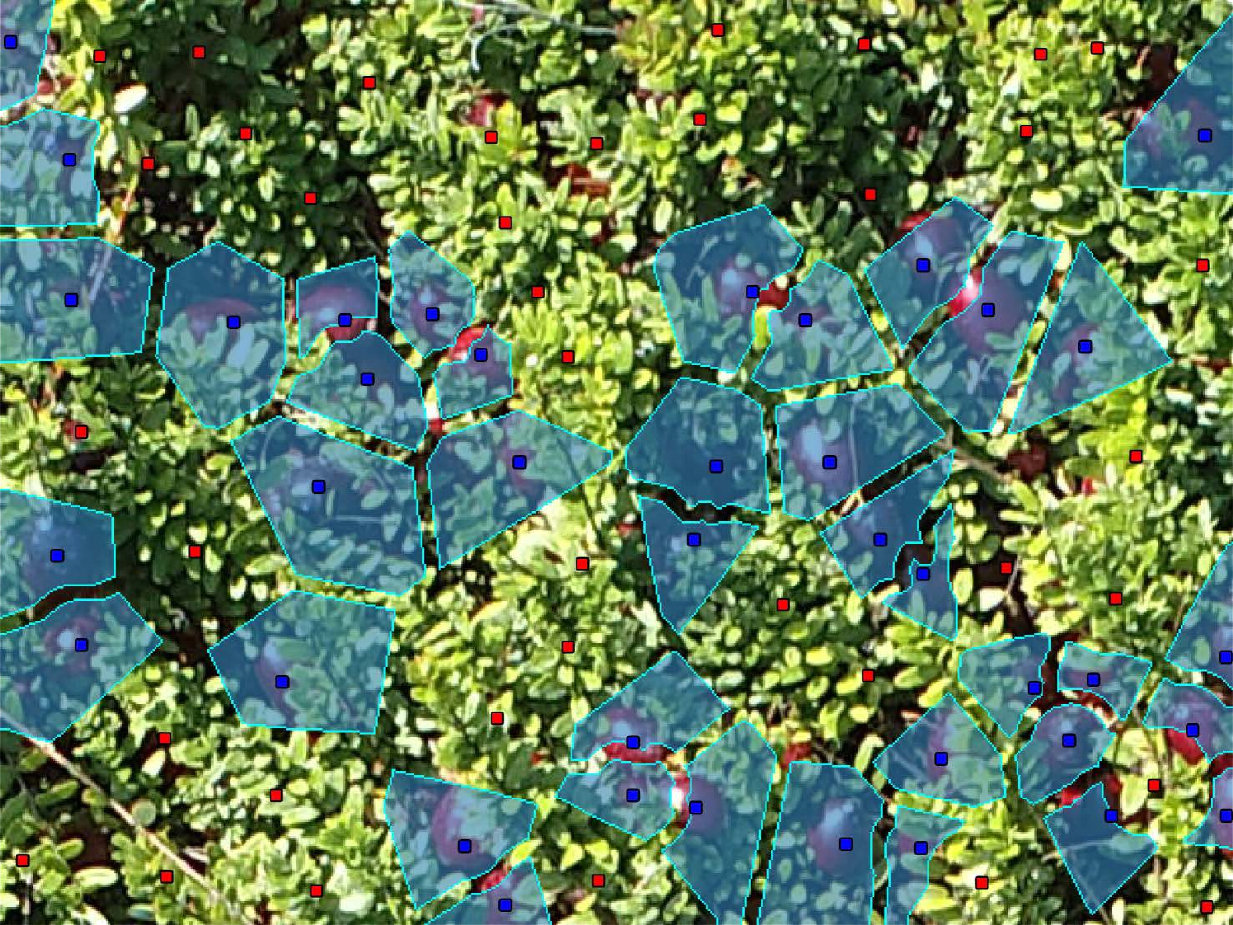}\\
    (a) & (b) & (c)\\
\end{tabularx}
\vspace{-0.35cm}
\caption{Visual comparison between (a) watershed \cite{meyer_1994}, (b) split watershed used in \cite{laradji2018blobs}, and (c) our selective watershed. Highlighted pixels are considered in their respective method. Our method utilizes negative ground truth points (also called background) to generate more selective information better suited for learning. Note that (b) only considers the set of pixels representing the borders generated by the split watershed. Blue and red markers represent positive and negative ground truth points.  Best viewed in color and zoomed.}
\vspace{-0.5cm}
\label{selectivewatershed}
\end{figure*}
%
%
\subsection{Annotation Procedure}
We annotate 21,436 images with center points for training, and 702 images with pixel-wise annotations for testing and evaluation. 
All annotations are peer reviewed by other annotators through consensus, a process in which a given annotated image is passed to at least one more annotator for further labeling before it is submitted for a final review.
\vspace{-0.45cm}
\paragraph{Center Point Annotations.} Annotators are instructed to locate and tag cranberry center points, and equal number of background points. Background points are annotated at random locations, as far as possible from nearby cranberry annotations. 
\vspace{-0.45cm}
\paragraph{Berry-wise Annotations.} Annotations follow two main guidelines: (1) only visible cranberries are annotated; (2) if a cranberry is occluded by leaves, the occluded parts are included  resulting in a pixel-wise annotation that captures the shape of the occluded cranberry, hence termed {\it berry-wise} annotations. While order of visibility is not preserved during this annotation procedure, the annotations are instance-wise, which allows separability of objects if needed.
%
\subsection{Dataset Details}
CRAID has an average of 39.22 cranberries per image, with minimum count of 0 and maximum count of 167. Berry-wise annotated images have an average of 33.72\% pixel cover. Average point-wise annotation time for a single image is 4.32 minutes, while berry-wise annotations take 22.13 minutes. Our relatively high annotation time compared to average estimated time reported in \cite{bearman2016s} is mainly caused by image complexity and high object counts.

\def\Laplace{\slantbox[-.45]{$\mathscr{L}$}}
\section{Triple-S Network}
%
Our approach is built upon U-Net \cite{ronneberger2015u} and consists of three branches: segmentation, split, and shape, constructing our proposed Triple-S Network illustrated in Figure~\ref{model}. The segmentation branch aims to provide overall segmentation loss against point ground truth. The split and shape branches separate and refine individual blobs in segmentation outputs in accordance to boundaries and shape priors. The overall loss function is defined by
%
\begin{equation}
    \begin{aligned}
        \mathcal{L}_{}(X,Y) = &\lambda_0\mathcal{L}_{\text{Seg}}(X,Y) +  \lambda_1\mathcal{L}_{\text{Split}}(X,Y) \\ & + \lambda_2\mathcal{L}_{\text{Shape}}(X,Y)\text{,}
    \end{aligned}
\end{equation}
where $X$ and $Y$ represent the set of input images and point annotations respectively, and $\lambda_{\ast}$ represents the weights of proposed losses.
We define  $y$ as the set of ground truth points (from berry-wise annotations) and $\Tilde{y}$ as the predicted mask of image $x$. 
\subsection{Segmentation Branch}
$\mathcal{L}_{Seg}$ aims to encourage the model towards correct blob localization guided by positive point annotations. Let  $\Tilde{s}$ be the softmax probability prediction output for image $x$. Let $p_n$, $p_p \in y$ denote the positive and negative ground truth points, respectively. We then define the segmentation loss by
%
\begin{equation}
    \mathcal{L}_{Seg}(\Tilde{s},y) = - \sum_{p_p} log(\Tilde{s}) - \sum_{p_n} (1-log(\Tilde{s})) \text{.}
\end{equation}
%
%
\vspace{-0.45cm}
\subsection{Split Branch}
The split loss function serves two purposes: discourage overlapping instances, and define expansion direction for predicted segments. 
We define the selective watershed algorithm $\text{W}_{\text{select}}$, a modification of $\text{W}$ \cite{meyer_1994}, to utilize both positive and negative markers to produce background and object specific regions.
Using $\text{W}(y)$, we obtain a set $R$ of distinct regions for all ground truth points.
We can then find the set of positive regions 
as $r_p = \{r \in R : p_p \cap r \neq \varnothing\}$. The set of negative regions $r_n$  is defined similarly and $r_p \cap r_n = \varnothing.$
Figure \ref{selectivewatershed} visualizes the set of pixels considered in our method compared to other variations of the watershed algorithm. We apply $\text{W}_{\text{select}}$ on $\Tilde{y}$ with $y$ as markers to obtain $r_p$, the set of pixels representing the regions each instance can expand to without stepping onto other instances. 
The set $r_p$ is then passed through an erosion algorithm \cite{haralick1987image} to better distinguish instances' boundaries. The complete split loss function is
\begin{equation}
    \begin{aligned}
    \mathcal{L}_{Split}(\Tilde{y},y)  =  -&\sum_{ r_p} \mathcal{E}(\text{W}_{select}(\Tilde{y},y))\\ 
    &-\sum_{ r_n} \mathcal{E}(\text{W}_{select}(\Tilde{y},y))\text{.}
    \end{aligned}
\end{equation}
%
$\text{W}_{\text{select}}$ represents the selective watershed algorithm, and $\mathcal{E}$ represents the erosion algorithm \cite{haralick1987image}.
%
\subsection{Shape Branch}
This work examines two shape priors appropriate for cranberries: convexity and circularity. 
These priors 
are used to guide the model towards meaningful structuring of the predicted blobs to fit berry-wise ground truth.  
\vspace{-0.35cm}
\paragraph{Convexity Loss.} Let $B$ represent the set of all blobs (distinct contiguous set of pixels)  in $y$ detected using the connected components algorithm \cite{dillencourt1992general} and let $b\in B$ denote an individual blob. Similarly, let $\Tilde{B}$ be the set of blobs detected in $\Tilde{y}$ with  $\Tilde{b}  \in \Tilde{B}$ denoting an individual blob.
We can define a convexity measure of a predicted blob as the ratio between the blob area  to its convex hull as follows
\begin{equation}
    \mathcal{C}(\Tilde{b}) = \frac{area(\Tilde{b})}{area(ConvexHull(\Tilde{b}))}\text{ .}
\end{equation}
%
Since objects in our dataset are always circular or 
elliptical
when accounting for occlusion, the area enclosed by the predicted blob should always match or almost match the area enclosed by its convex hull, meaning our convexity measurement for each ground truth blob $b \in B$ is always close to one. The convexity loss $\mathcal{L}_{Convex}$ general form is given by
\begin{equation}
    \begin{aligned}
        \mathcal{L}_{Convex}(\Tilde{y},y) = &  \frac{1}{|\Tilde{B}|}\sum_{\Tilde{b} \in \Tilde{B} \in \Tilde{y},b \in B \in y} z(\Tilde{b},b)\text{,}
    \end{aligned}
\end{equation}
where 
\begin{equation}
    \begin{aligned}
        z(\Tilde{b},b) = 
        \begin{cases}
        \frac{1}{2}(\mathcal{C}(\Tilde{b}) - \mathcal{C}(b))^{2}, \\
        \indent \indent \text{if } |\mathcal{C}(\Tilde{b}) - \mathcal{C}(b)| < 1\\
        |\mathcal{C}(\Tilde{b}) - \mathcal{C}(b)| - \frac{1}{2}, \text{ otherwise.}
        \end{cases}
    \end{aligned}   
\end{equation}
This is a variation of Huber loss \cite{huber1992robust}, in which the gradient is calculated with respect to the residual
(supplementary for details).
Note that $|\Tilde{B}|$ represents the cardinality of that set. 
\vspace{-0.35cm}
\paragraph{Circularity Loss.} Another approach is to directly find the circularity difference between the predicted blob $\Tilde{b} \in \Tilde{B}$ and ground truth $b \in B$. 
We formulate our loss function similar to the unconstrained nonlinear programming problem formulation proposed by \cite{murthy1980minimum}, looking to find the least square reference circle (notated as $LSC$) of a given blob. Let $r_b$ be the set of radii originating at the center of blob $b$. The circularity measurement of blob $b$ is then defined by the difference between the maximum and minimum radii that exist in $r_b$. The reference circle LSC is given by
\vspace{-0.03cm}
\begin{equation}
    \begin{aligned}
        LSC(b) &=\max(\sqrt{(u_i - u_c)^2 + (v_i-v_c)^2})\\
        &- \min(\sqrt{(u_i - u_c)^2 + (v_i-v_c)^2}) \forall i \in b\text{,}\\
    \end{aligned}
\end{equation}
\vspace{-0.03cm}
where $(u_i,v_i)$ are the coordinates of pixel $i$ in blob $b$ with center $(u_c,v_c)$. Coordinates are with respect to the input image $x$. The final circularity loss formulation is defined as
\begin{equation}
    \begin{aligned}
        \mathcal{L}_{Circ}(\Tilde{y},y) = &  \frac{1}{|\Tilde{B}|}\sum_{\Tilde{b} \in \Tilde{B} \in \Tilde{y},b \in B \in y} z(\Tilde{b},b)\text{,}
    \end{aligned}
\end{equation}
\vspace{-0.03cm}
with
%
\begin{equation}
    \begin{aligned}
        z(\Tilde{b},b) = 
        \begin{cases}
        \frac{1}{2}(LSC(\Tilde{b}) - LSC(b))^{2},\\
        \indent \indent \text{if } |LSC(\Tilde{b}) - LSC(b)| < 1\\
        |LSC(\Tilde{b}) - LSC(b)| - \frac{1}{2}, \text{ otherwise.}
        \end{cases}
    \end{aligned}
\end{equation}
Since we want to predict circular or close to circular blobs, our ground truth $LSC(b)$ is zero, encouraging minimal difference between maximum and minimum radii.

\subsection{Count Branch}
This section explains the count loss used for ablation study, in which we explore the contribution and advantage of direct count learning with and without the usage of shape priors. $\mathcal{L}_{Count}$ aims to directly guide the model towards the correct number of instances, $c$, present in image $x$. This branch first separates blobs $B$ from the segmentation prediction using connected components algorithm \cite{dillencourt1992general} noted as $CC$, and the resulting connected components count $\Tilde{c}$ is used as count prediction. This means that small regions present in the segmentation prediction results high count prediction, which penalizes the model and discourages it from such predictions. More formally
%
%
\begin{equation}
    \begin{aligned}
        \mathcal{L}_{Count}(\Tilde{C},C) = &  \frac{1}{|C|}\sum_{c \in C, \Tilde{c} \in \Tilde{C}} z_c \text{ ,}
    \end{aligned}
\end{equation}
where
\begin{equation}
    \begin{aligned}
        z_c = 
        \begin{cases}
        \frac{1}{2}(\Tilde{c} - c)^{2},&\text{if } |\Tilde{c} - c| < 1\\
        |\Tilde{c} - c| - \frac{1}{2}, & \text{otherwise,}
        \end{cases} \\
    \end{aligned}
\end{equation}
Here, $\tilde{c} = |CC(\tilde{y})|$. $C$ and $\tilde{C}$ are the ground truth and predicted counts for all samples. $\mathcal{L}_{count}$ is small if the difference between $\Tilde{c}$ and $c$ is small. 
%
\begin{table*}[t!]
\centering
\large
    \begin{tabular}{ccccc}
     \toprule
     Training Ground Truth & Method $\backslash$ Metric & mIoU (\%) & MAE & $Q_{cs}$ \\
     \midrule
     Pixel-wise & U-Net \cite{ronneberger2015u} & 78.56 & 9.60 & 8.19\\
     \midrule
      \multirow{4}{*}{{Points}}& U-Net  \cite{ronneberger2015u} & 60.61 & 18.67 & 3.25 \\
      & \cite{ribera2019locating} & - & 39.22 & - \\
      & m\cite{ribera2019locating} & - & 21.76 & -\\
      & LC-FCN \cite{laradji2018blobs} & 61.97 & 17.46 & 3.55\\
     \midrule
     \multirow{7}{*}{{Points}} & Ours ($\mathcal{L}_{Seg}$ + $\mathcal{L}_{Split}$) & 67.39 & 16.33 & 4.13\\
      & Ours ($\mathcal{L}_{Seg}$ + $\mathcal{L}_{Split}$ + $\mathcal{L}_{Count}$) & 62.54 & \textbf{13.46} & 4.65 \\
      & Ours ($\mathcal{L}_{Seg}$ + $\mathcal{L}_{Split}$ + $\mathcal{L}_{Circ}$) & 67.85 & 14.25 & \textbf{4.76} \\
      & Ours ($\mathcal{L}_{Seg}$ + $\mathcal{L}_{Split}$ +  $\mathcal{L}_{Circ}$ + $\mathcal{L}_{Count}$) & 65.89  & 14.31 & 4.60\\
      & Ours ($\mathcal{L}_{Seg}$ + $\mathcal{L}_{Split}$ + $\mathcal{L}_{Convex}$) & \textbf{68.71} & 15.90 & 4.32\\
      & Ours ($\mathcal{L}_{Seg}$ + $\mathcal{L}_{Split}$ +  $\mathcal{L}_{Convex}$ + $\mathcal{L}_{Count}$) & 65.35 & 15.93 & 4.10\\
     \bottomrule
\end{tabular}
\caption{Mean Intersection over Union (\%) accuracy (higher is better), Mean Average Error (MAE) (lower is better), and Inverse MAE to mIoU ratio $\text{Q}_{\text{cs}}$ (higher is better) metrics on CRAID. Our proposed method outperforms the SOTA (trained with point annotations) in all metrics. All evaluation is done against pixel-wise ground truth.}
\vspace{-0.35cm}
\label{allresults}
\end{table*}

\vspace{-0.12cm}
\section{Experiments}
\begin{figure*}[t!]
\setlength\tabcolsep{1pt}
\def\arraystretch{0.5}
\centering
\begin{tabularx}{\textwidth}{cccc}
    \includegraphics[width=0.25\linewidth]{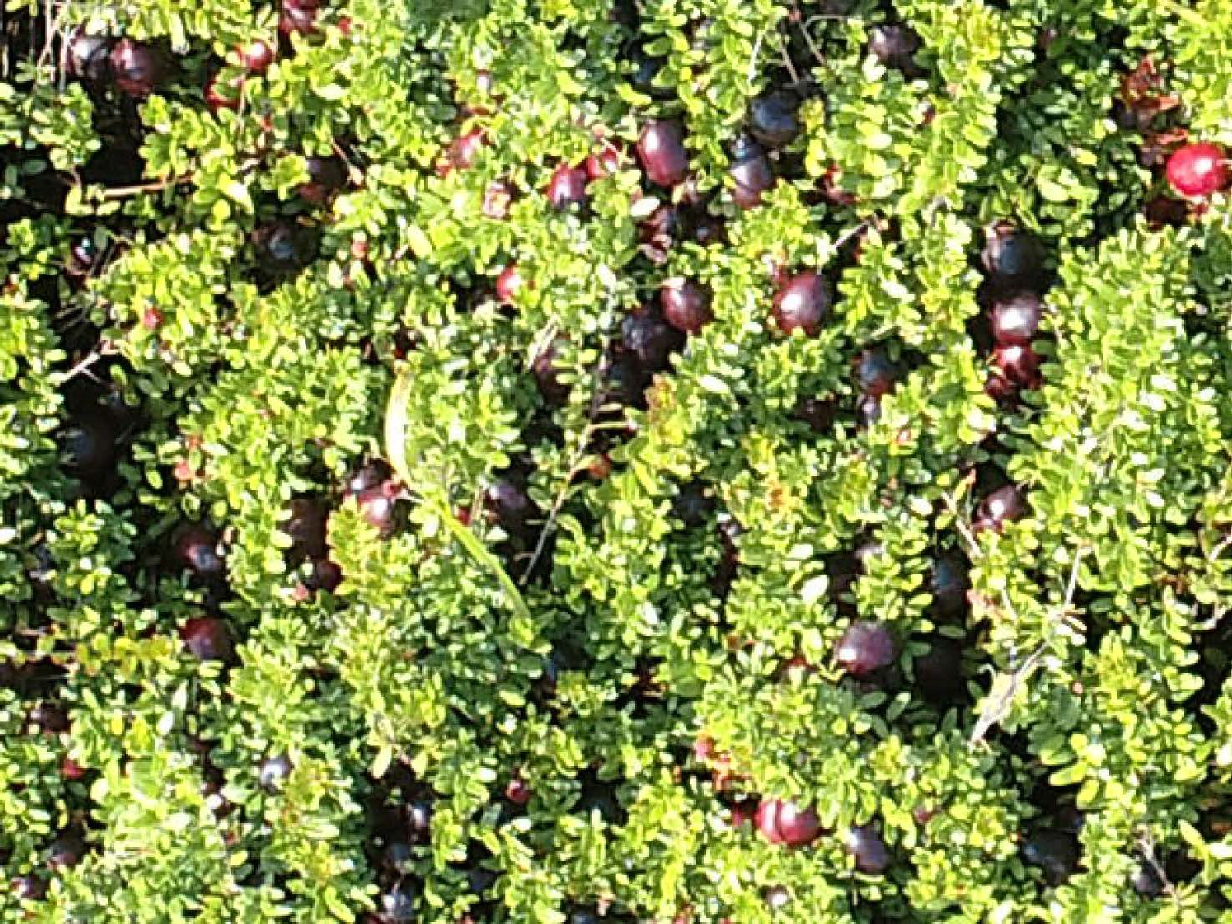} &   \includegraphics[width=0.25\linewidth]{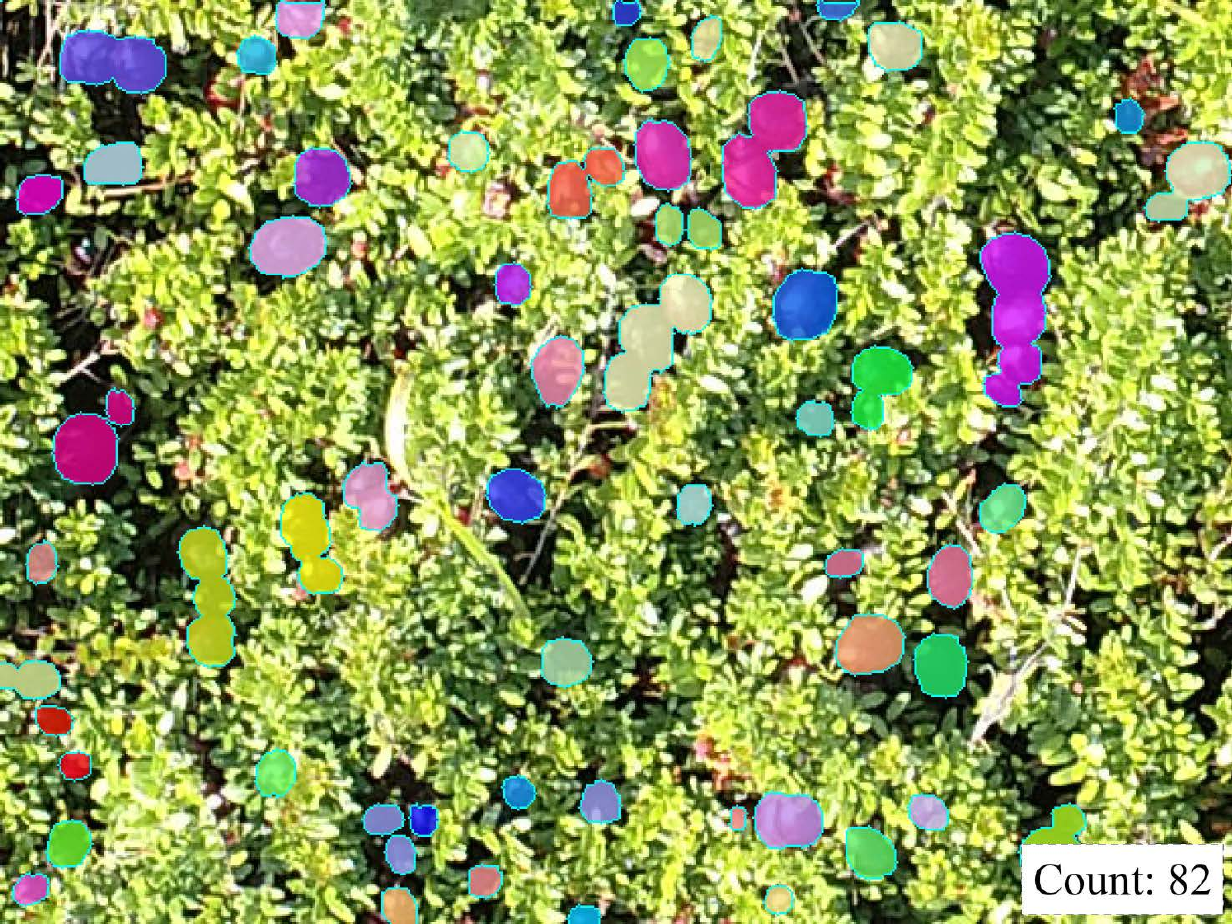} &   \includegraphics[width=0.25\linewidth]{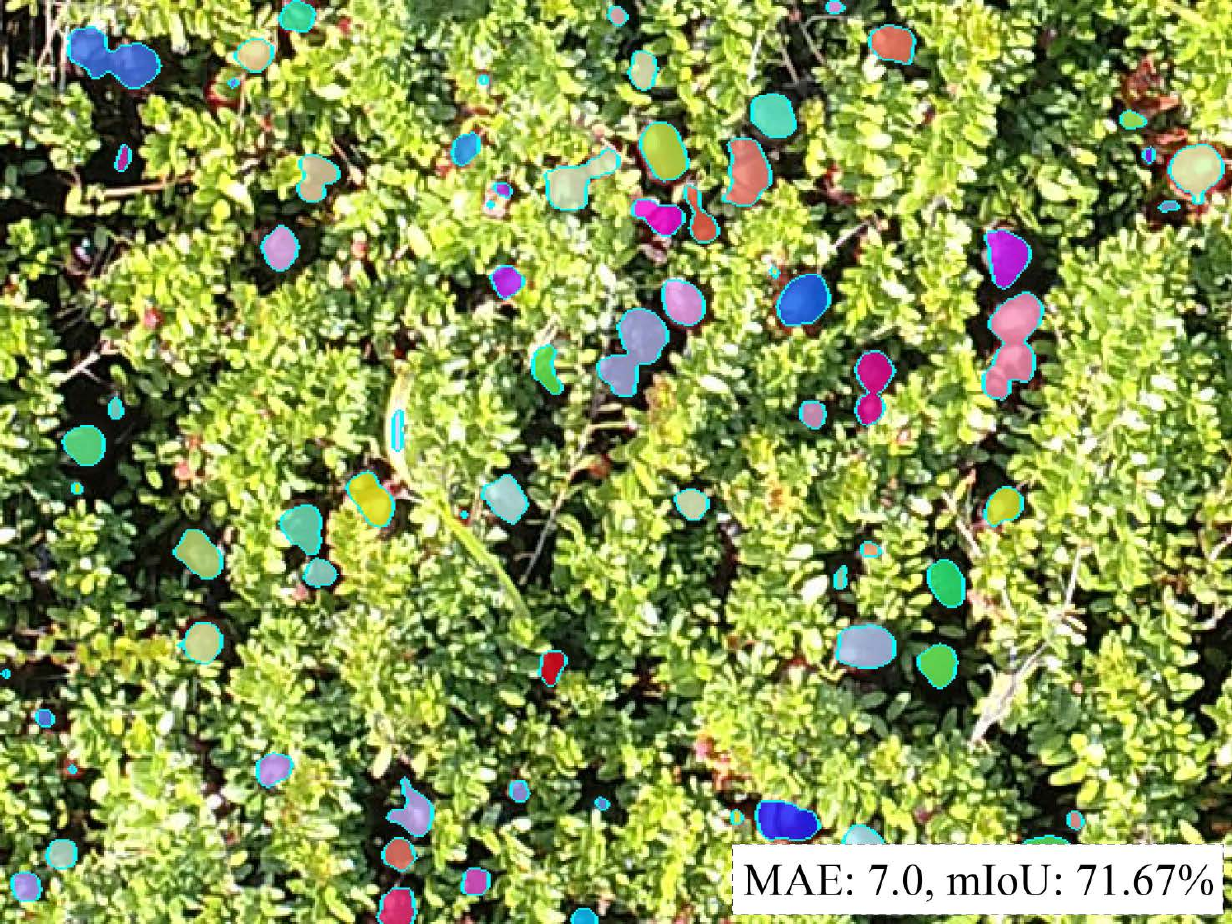} &   \includegraphics[width=0.25\linewidth]{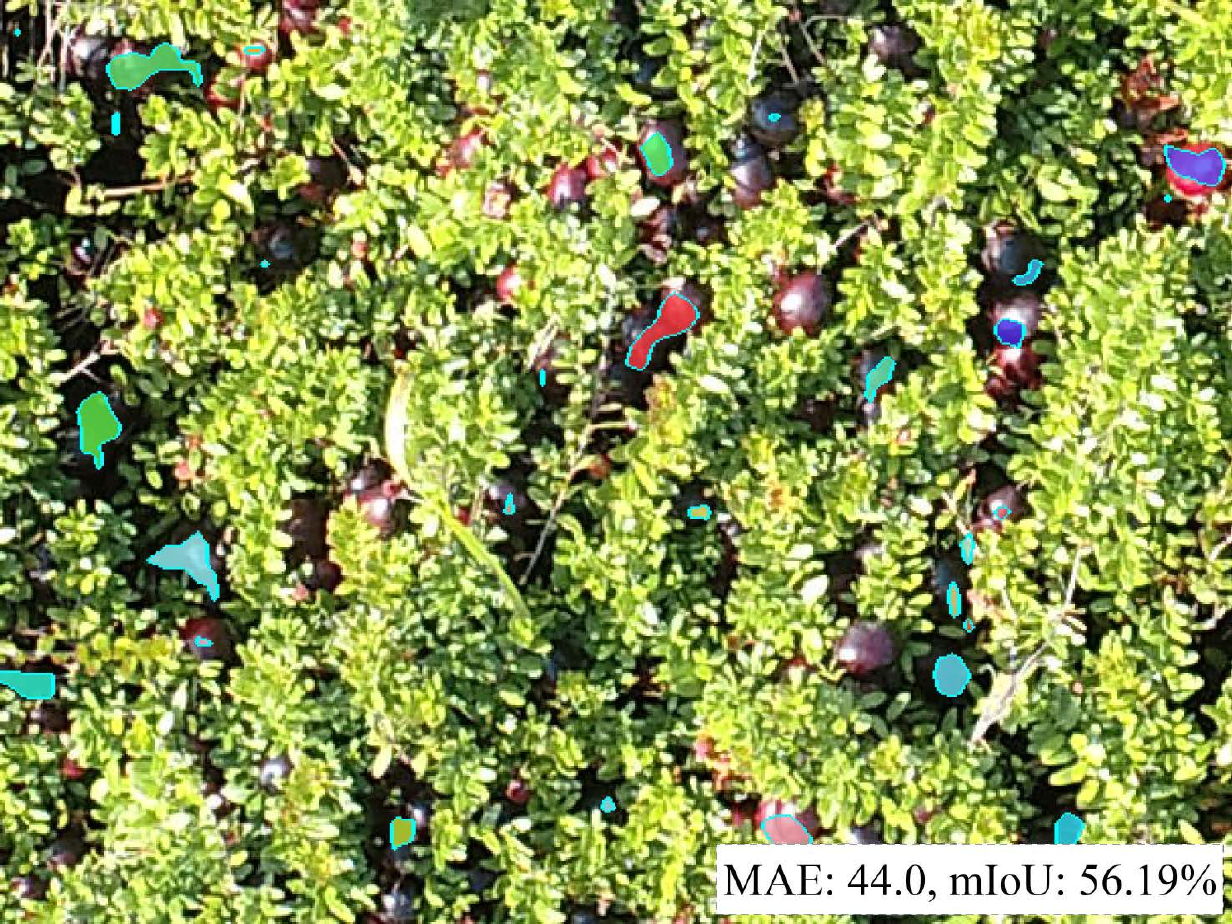} \\
    \includegraphics[width=0.25\linewidth]{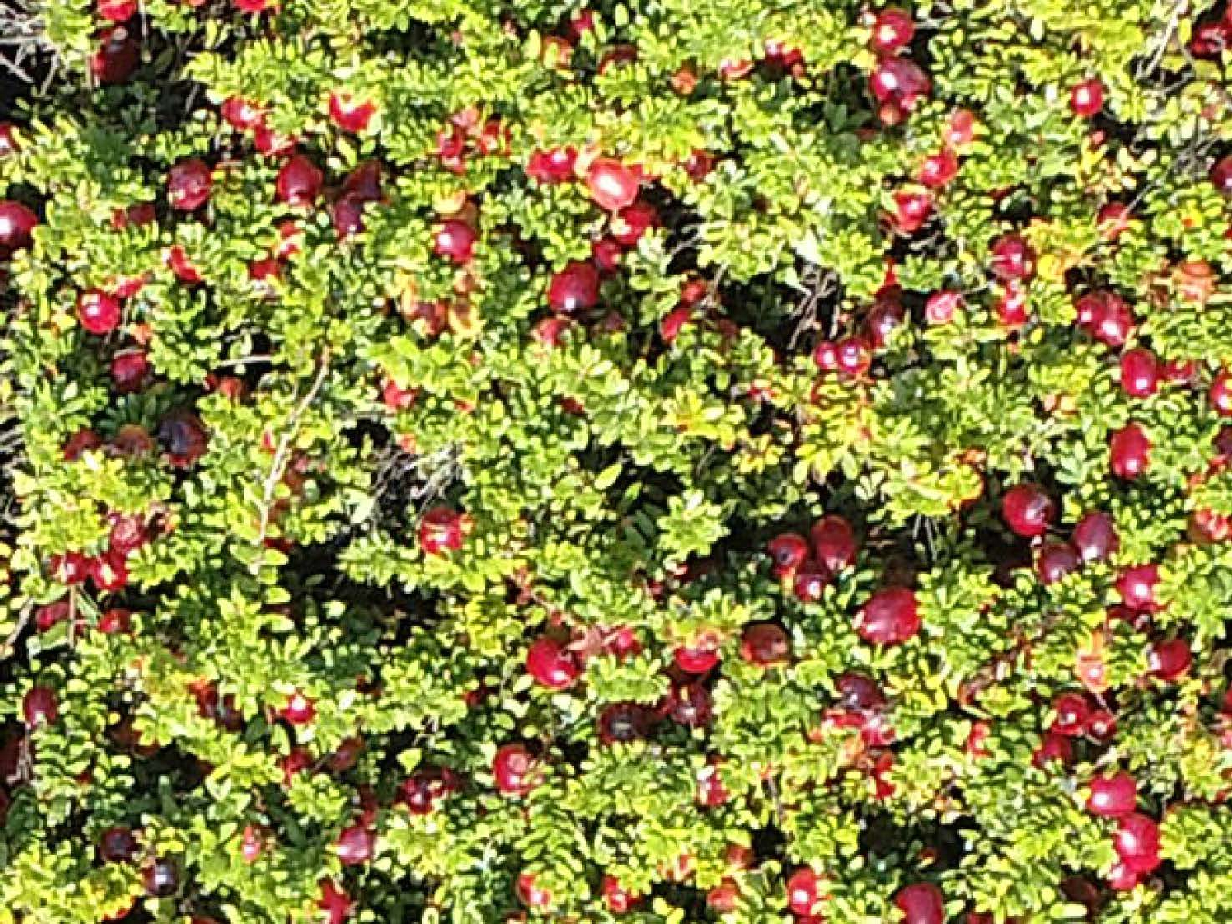} & \includegraphics[width=0.25\linewidth]{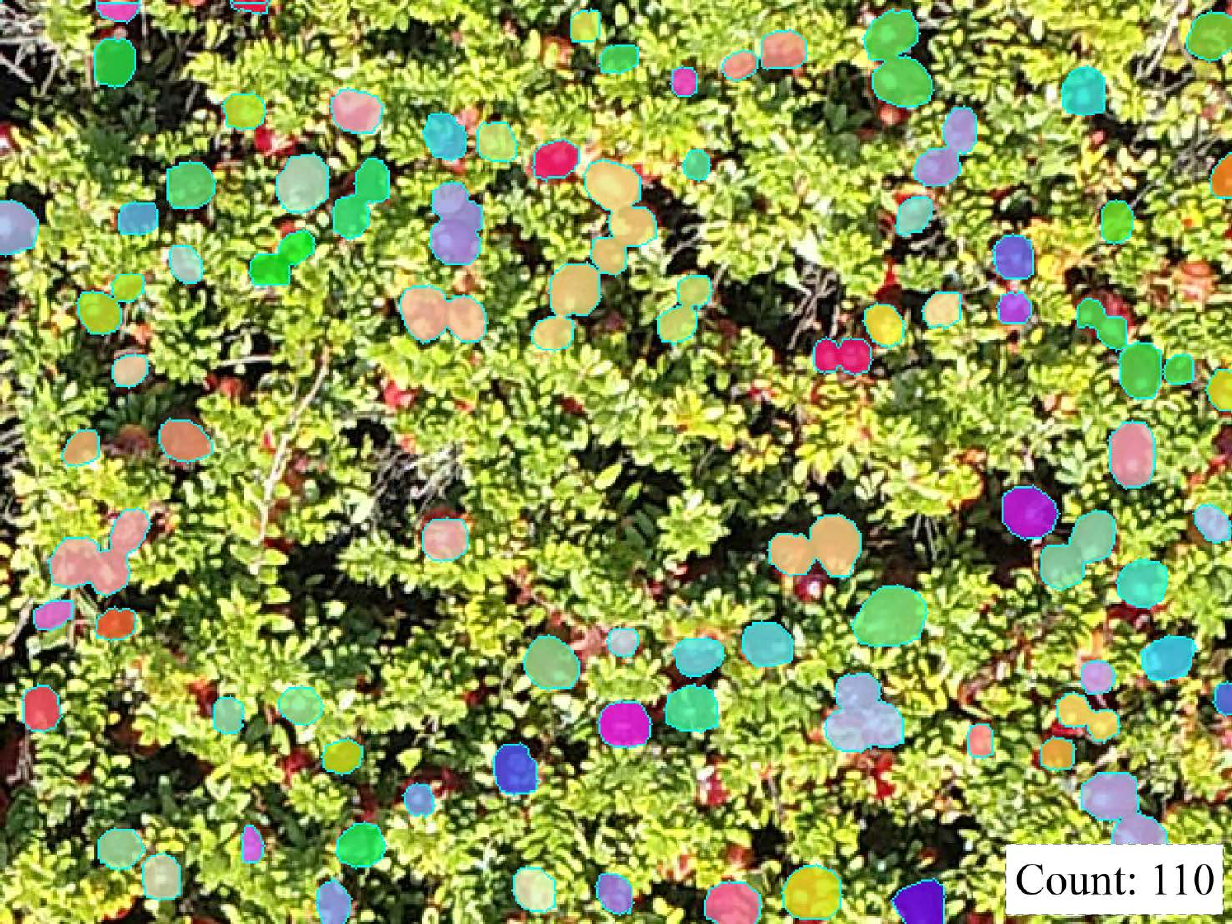} &   \includegraphics[width=0.25\linewidth]{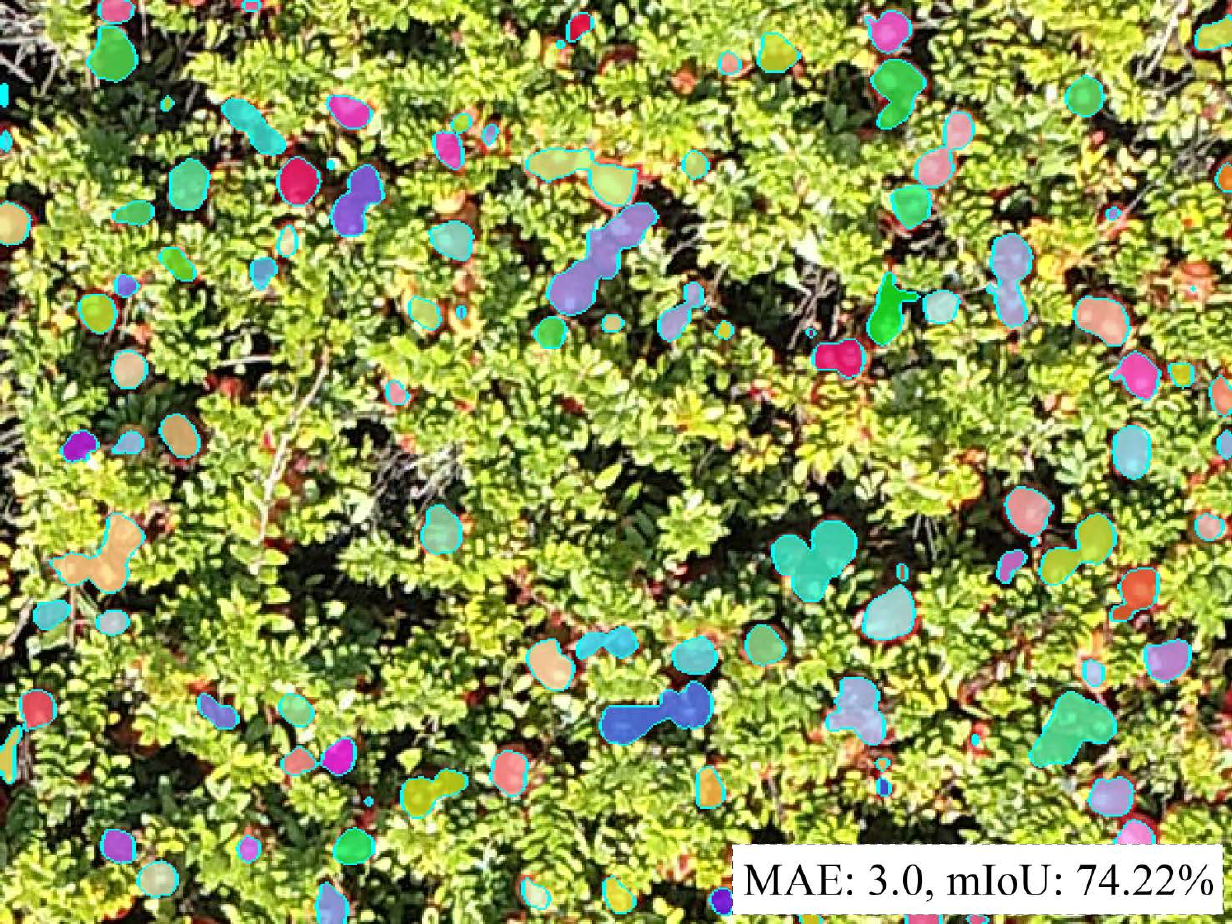} &   \includegraphics[width=0.25\linewidth]{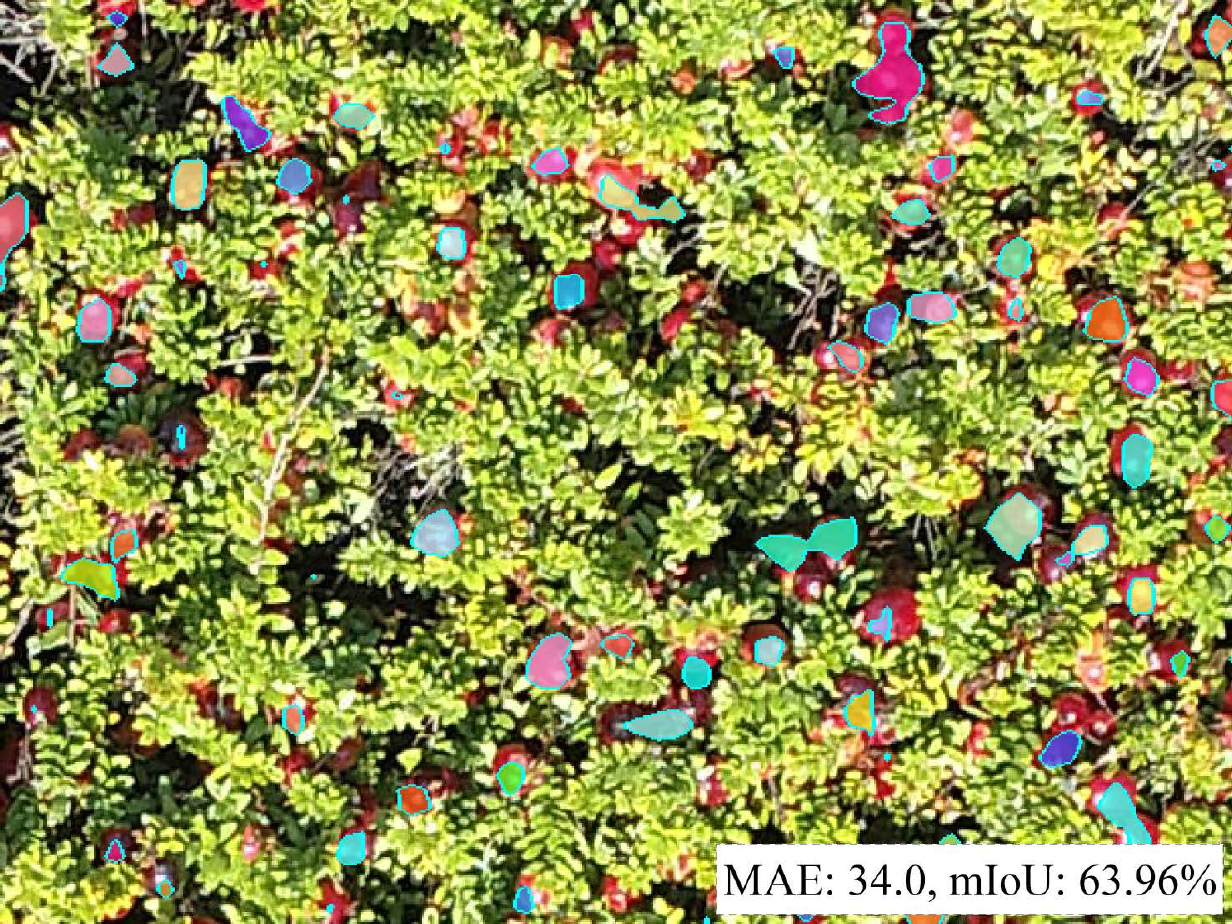} \\
    \includegraphics[width=0.25\linewidth]{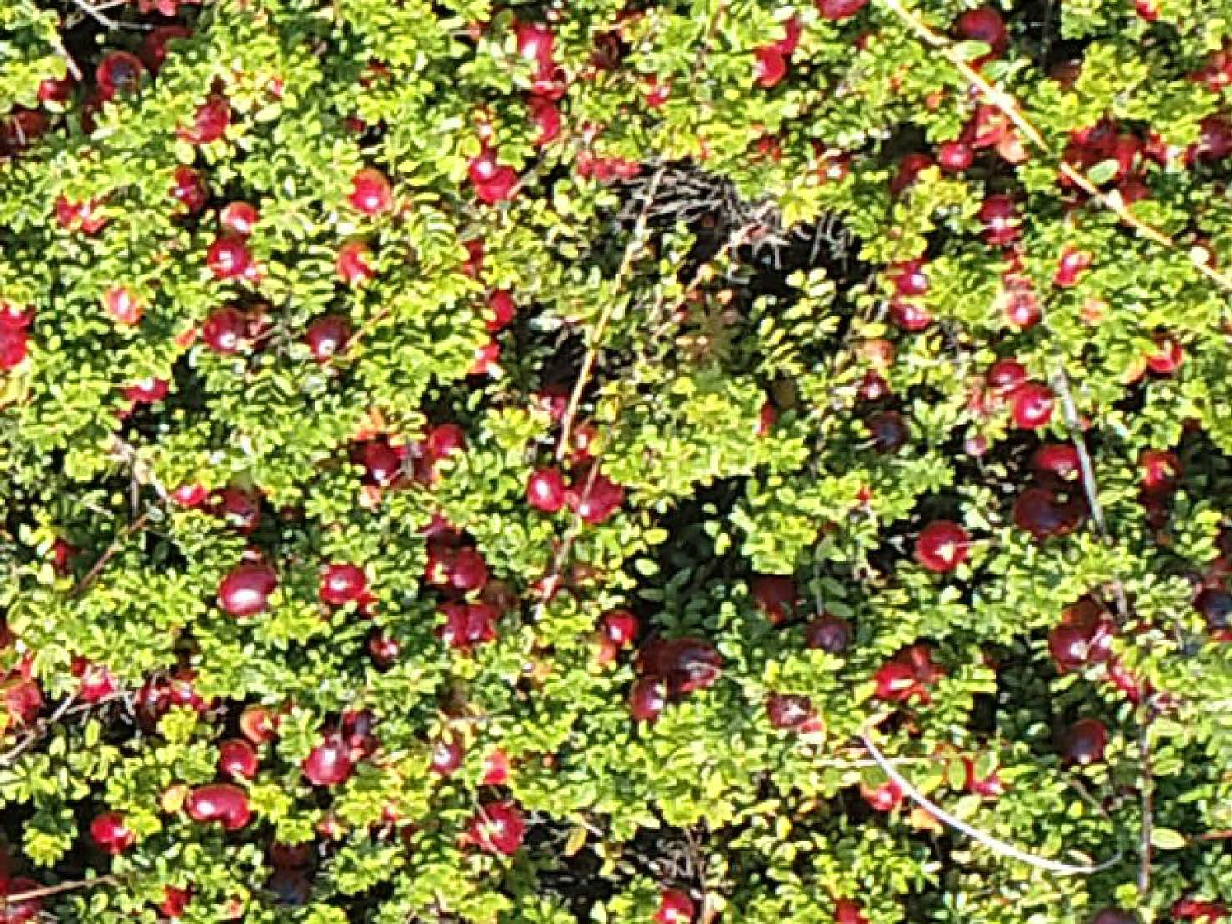} & \includegraphics[width=0.25\linewidth]{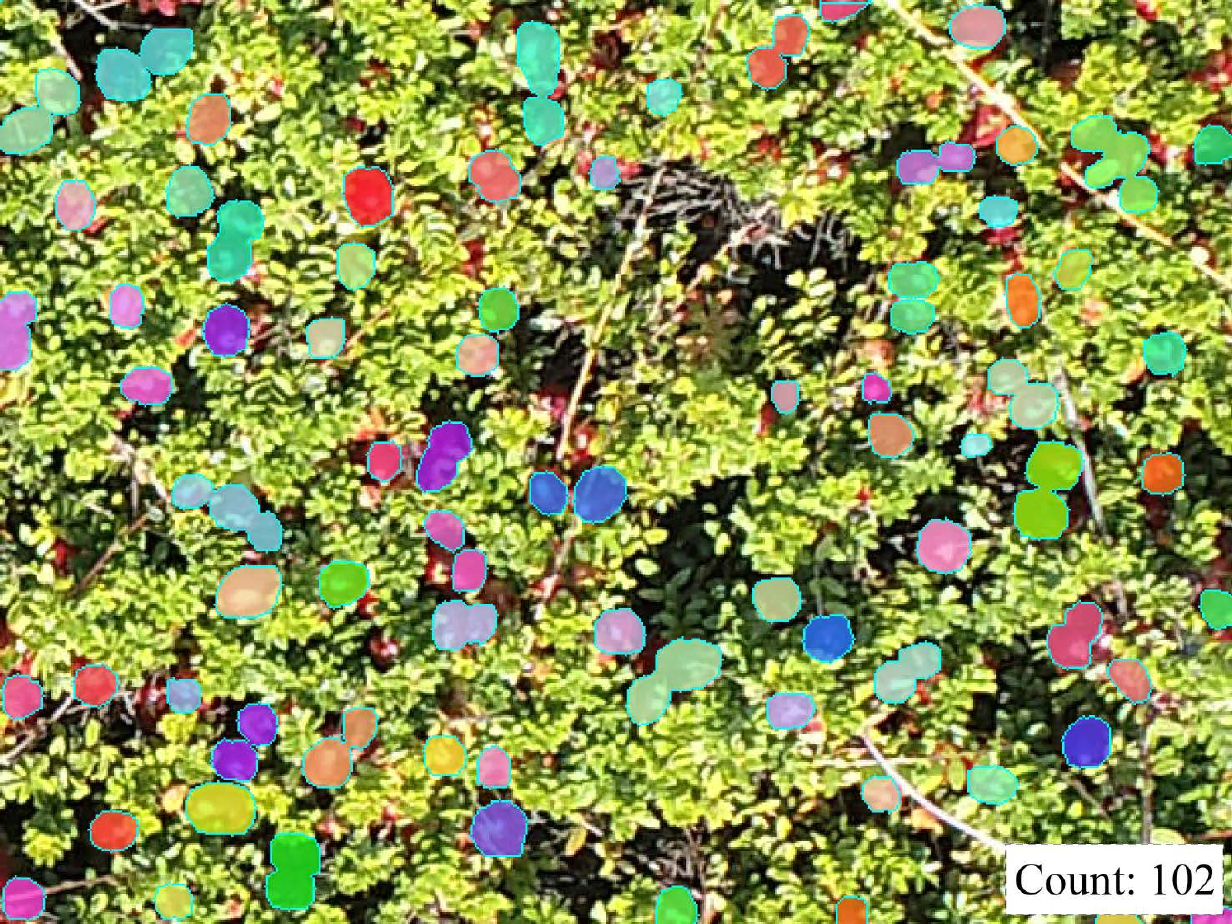} &   \includegraphics[width=0.25\linewidth]{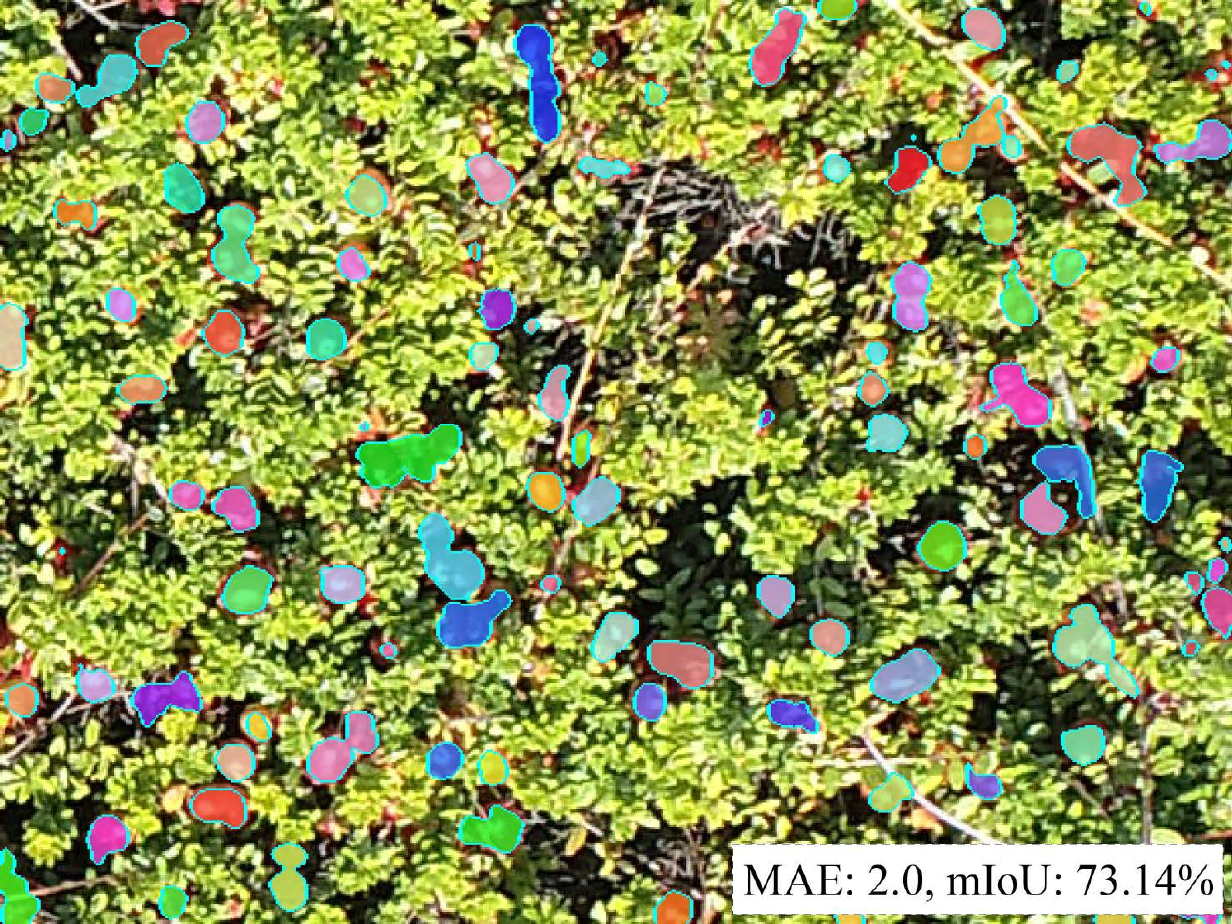} &   \includegraphics[width=0.25\linewidth]{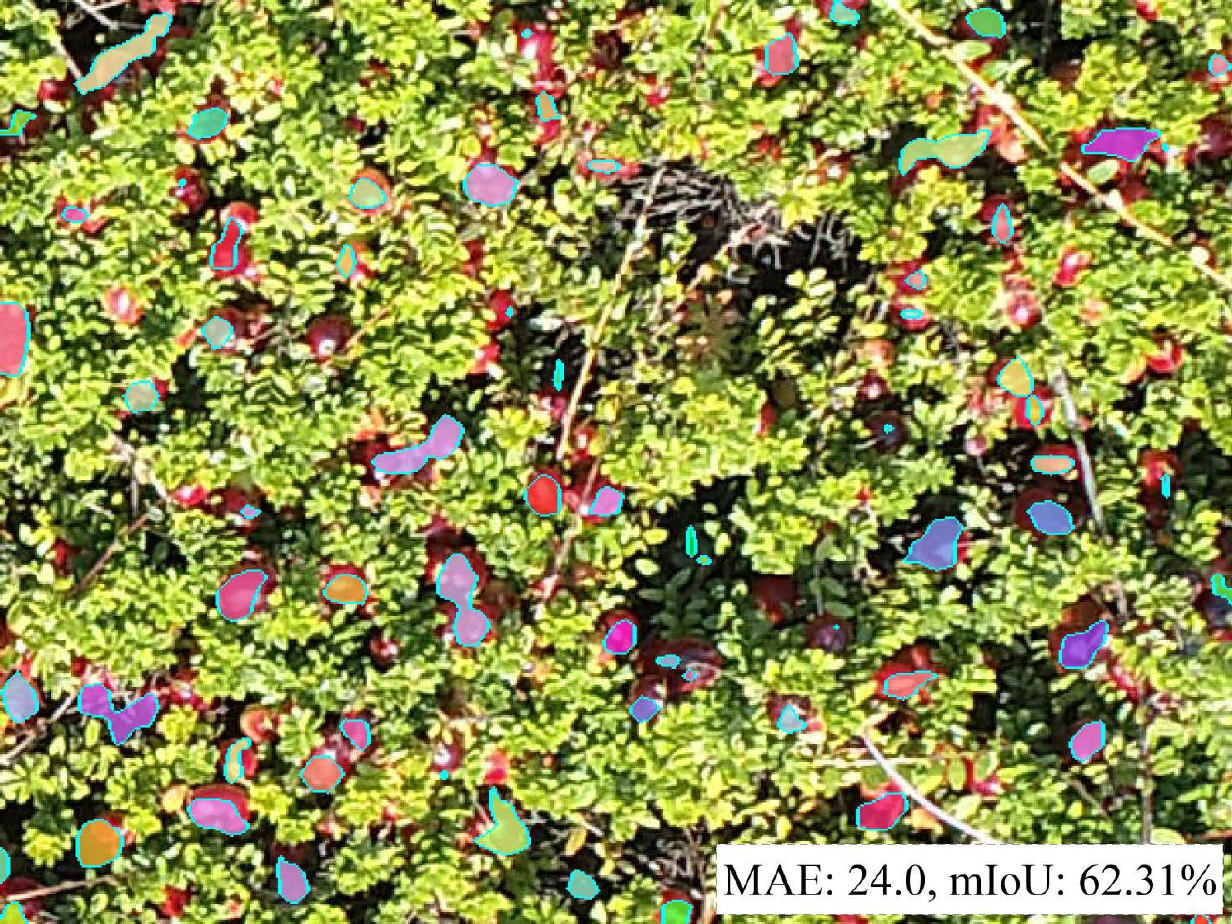} \\
    \includegraphics[width=0.25\linewidth]{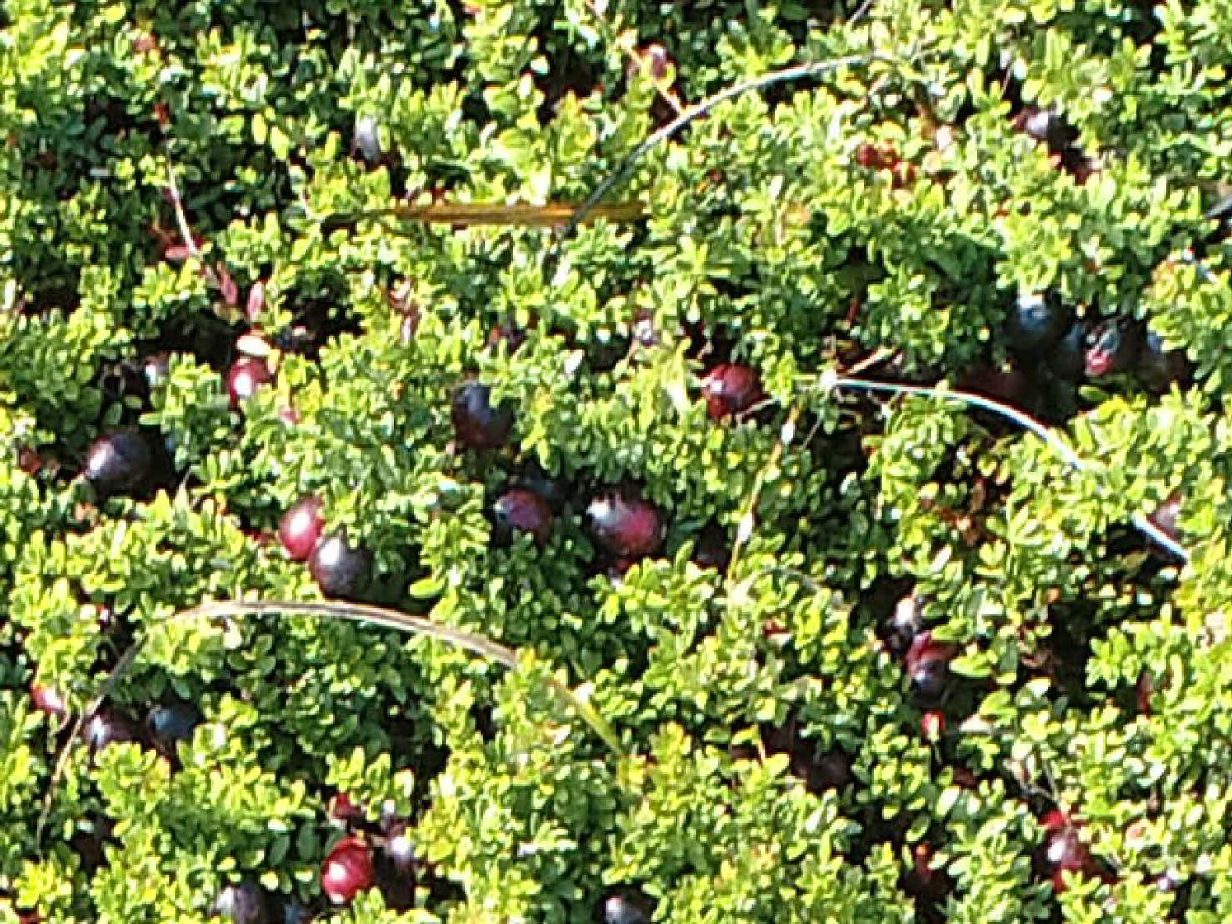} & \includegraphics[width=0.25\linewidth]{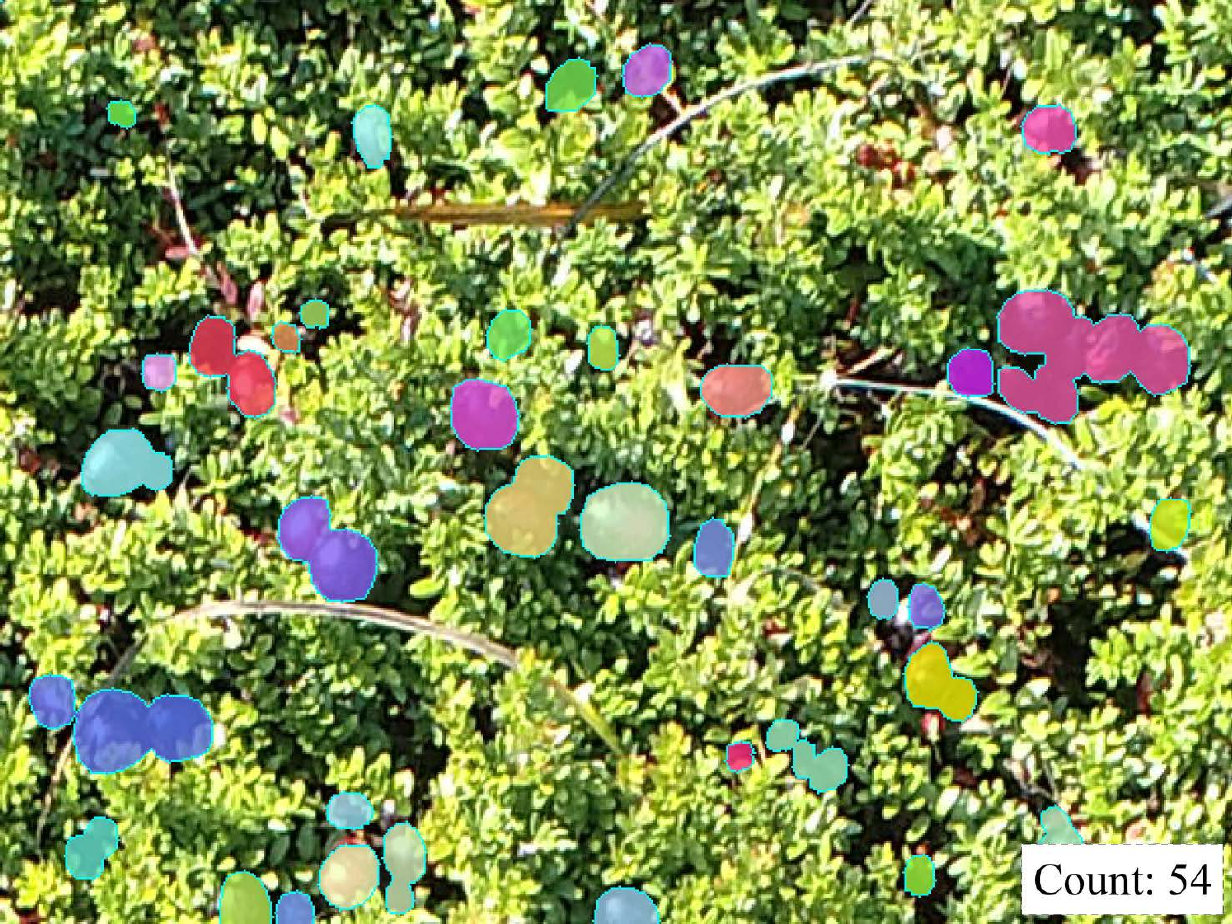} &   \includegraphics[width=0.25\linewidth]{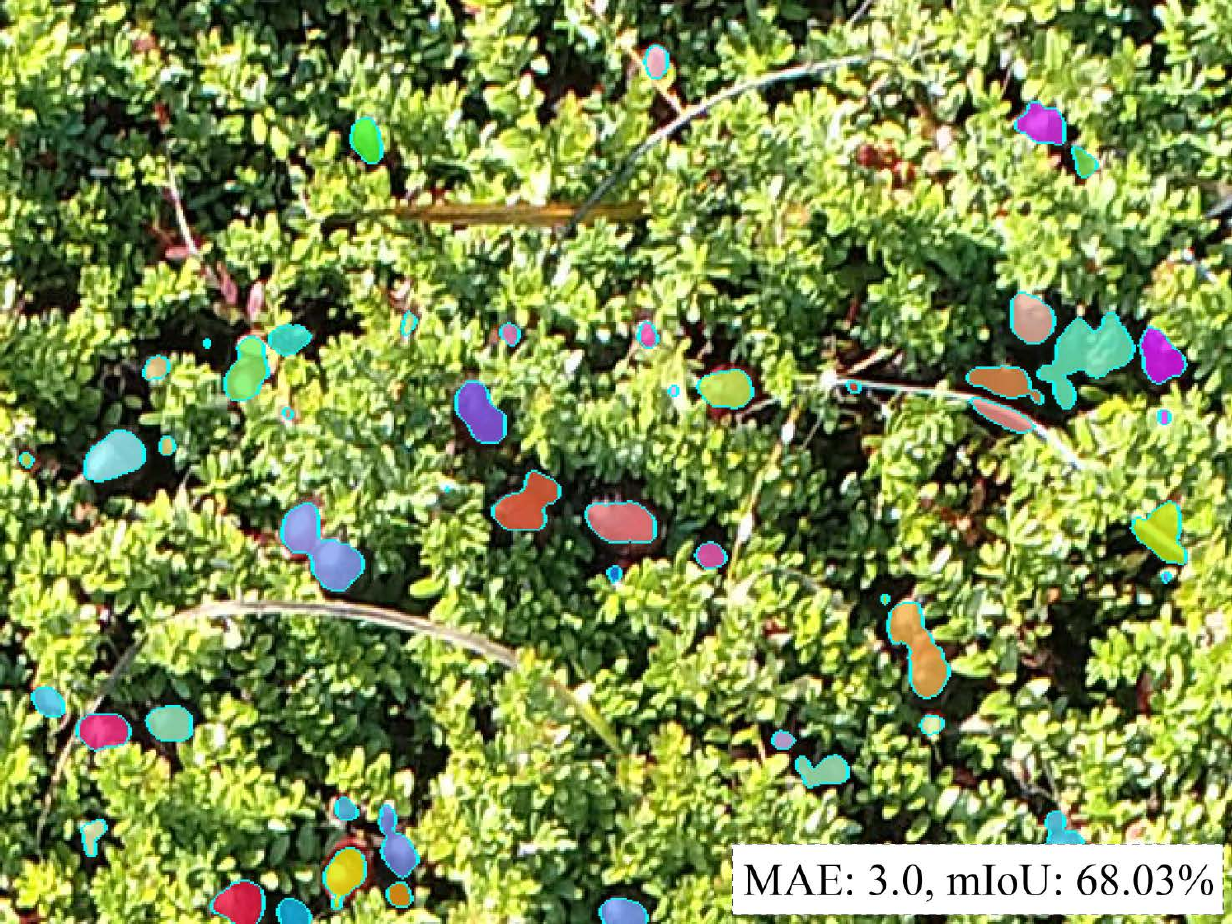} &   \includegraphics[width=0.25\linewidth]{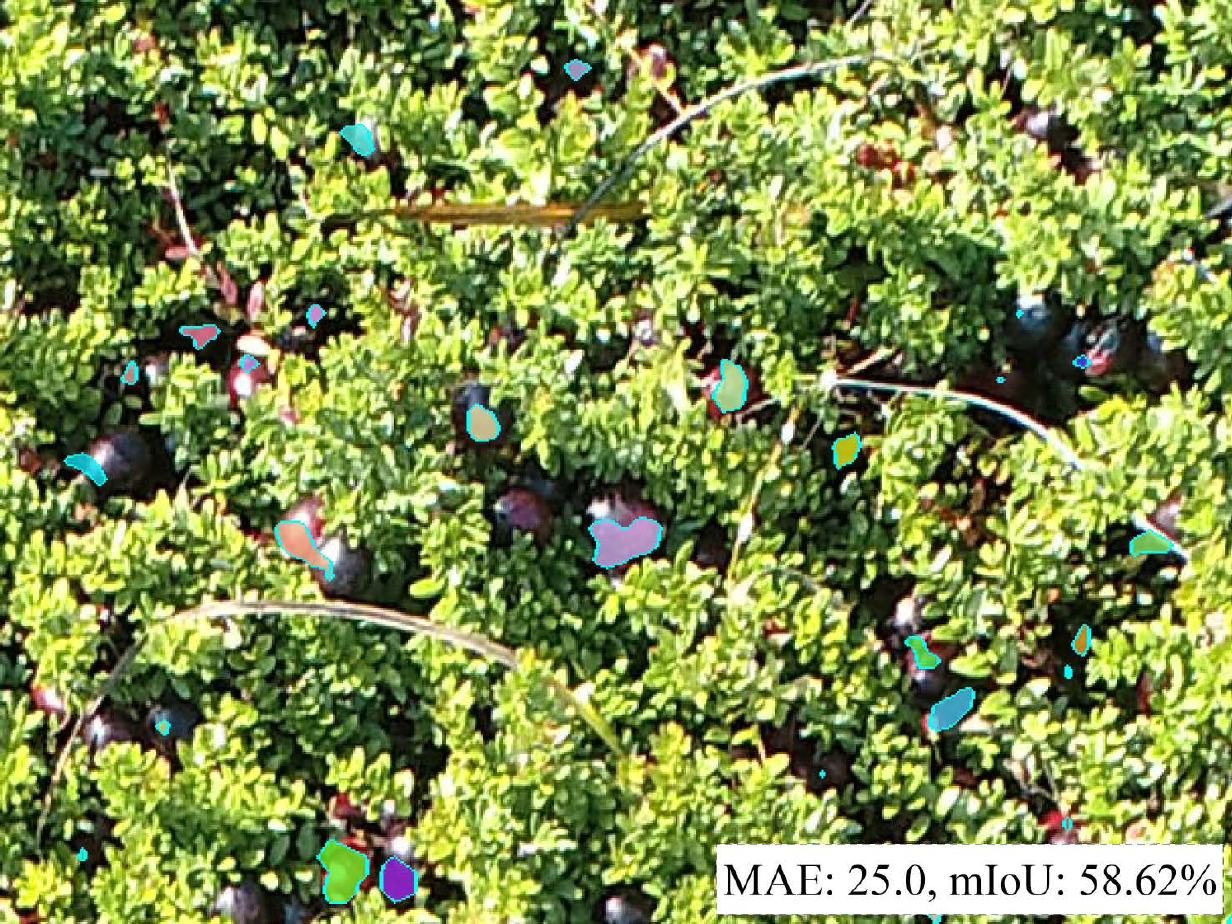} \\
    \includegraphics[width=0.25\linewidth]{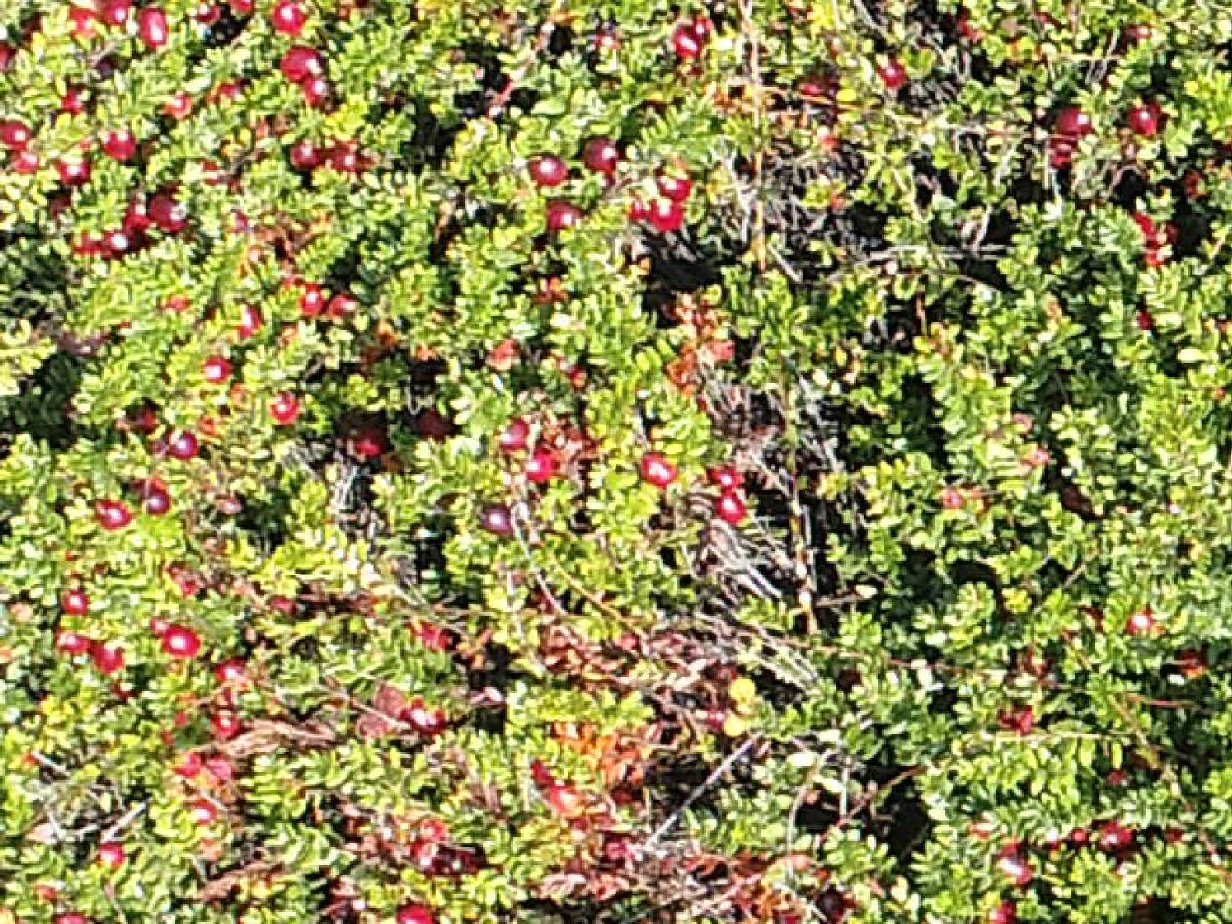} & \includegraphics[width=0.25\linewidth]{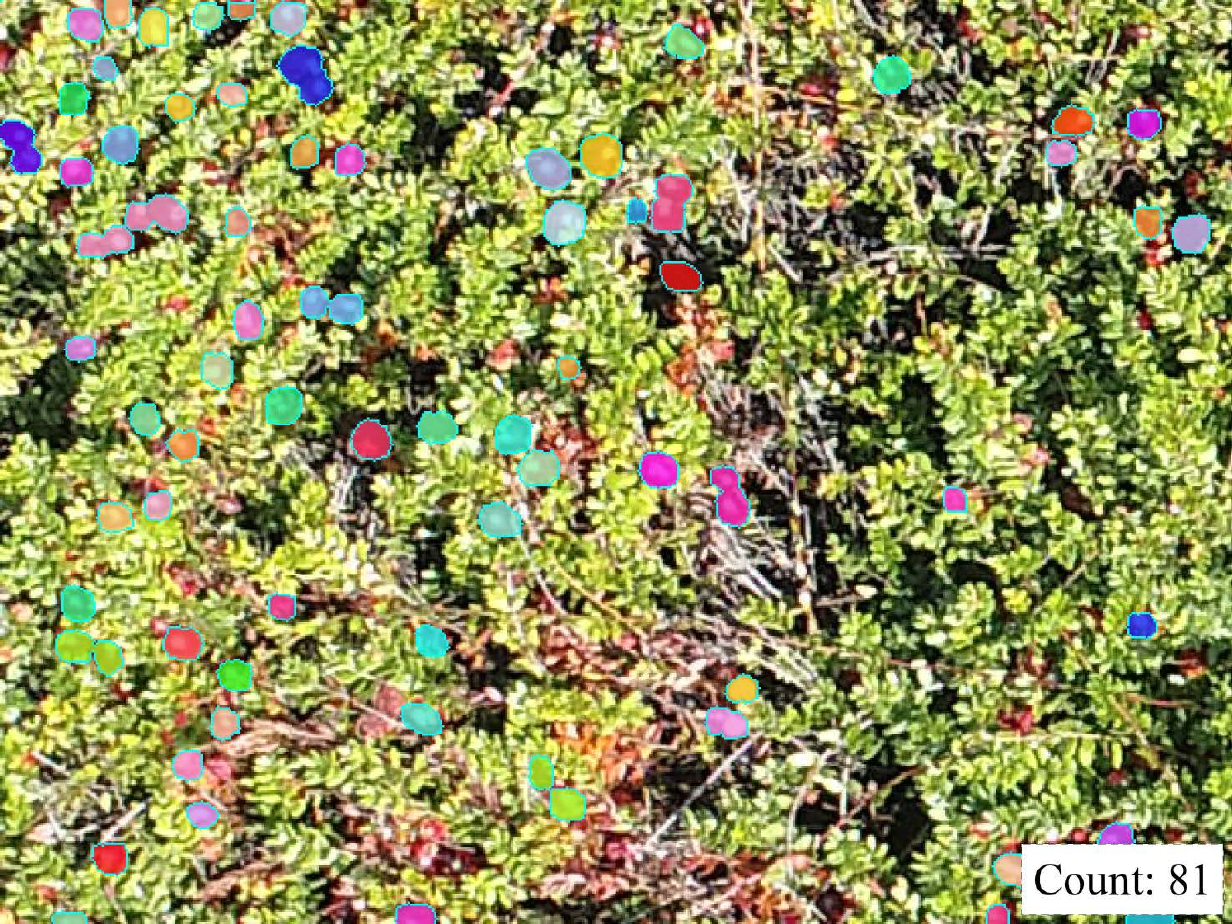} &   \includegraphics[width=0.25\linewidth]{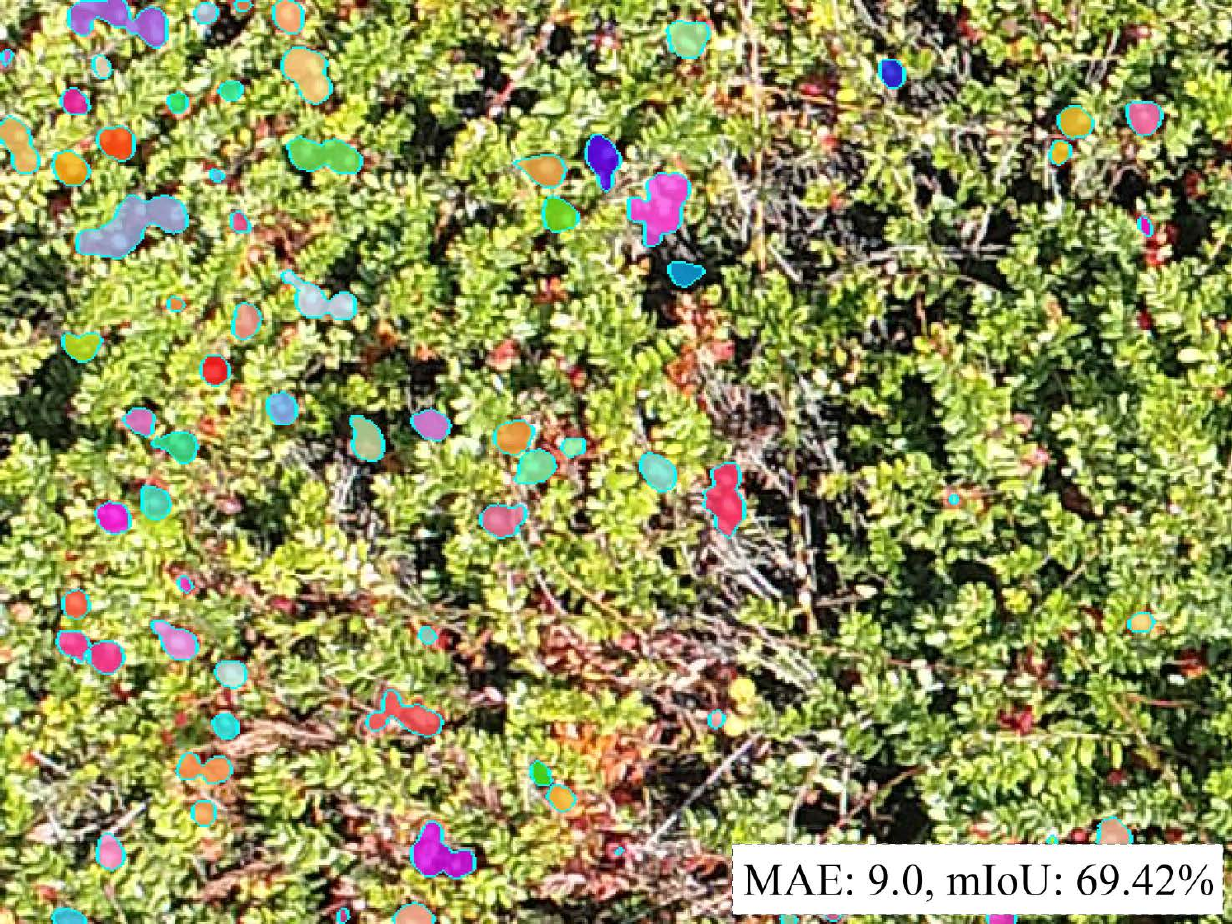} &   \includegraphics[width=0.25\linewidth]{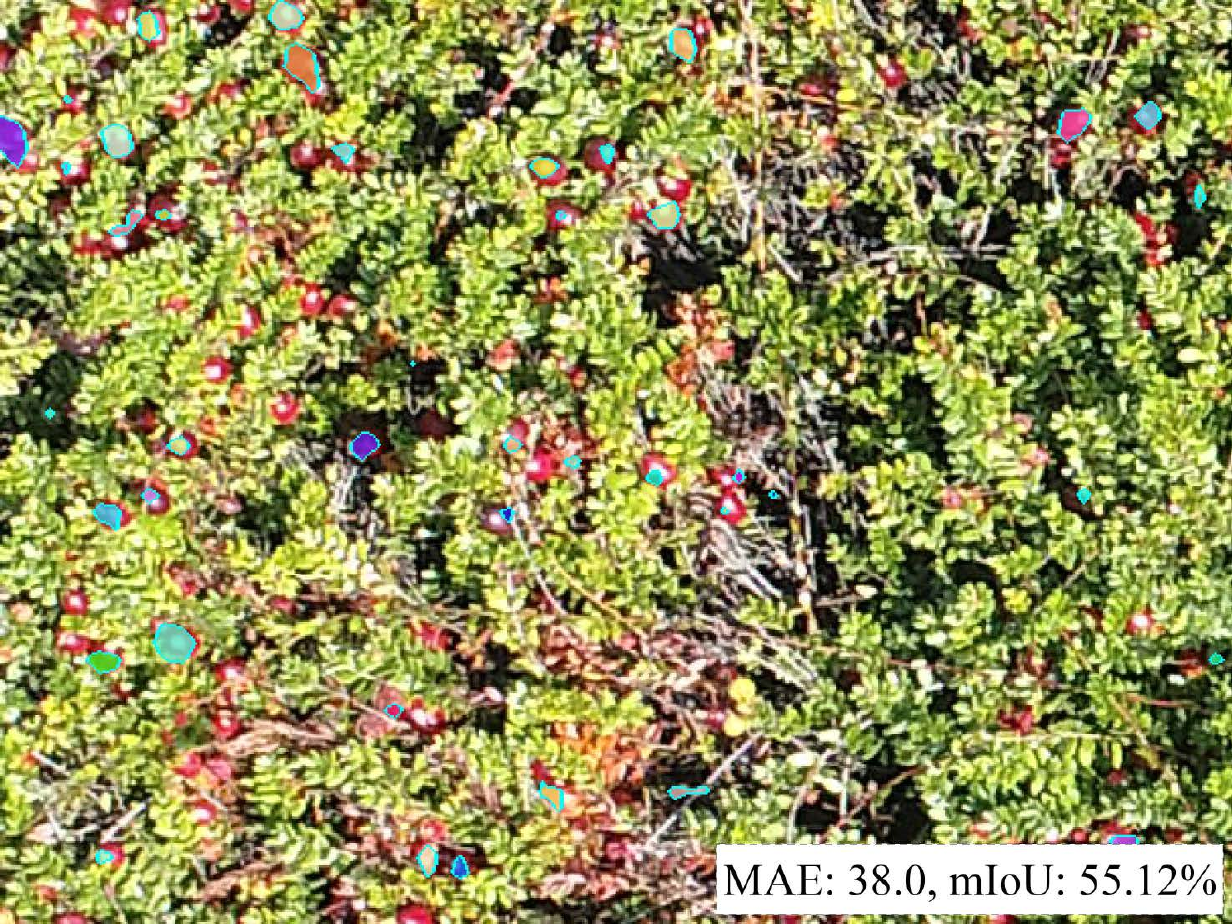} \\
    Input Image & Ground Truth & Ours & LC-FCN
\end{tabularx}
\vspace{-0.25cm}
\caption{Qualitative comparison with SOTA methods on CRAID. Our method ($\mathcal{L}_{Seg}$ + $\mathcal{L}_{Split}$ + $\mathcal{L}_{Convex}$) shows that using shape priors and better boundary and region selection allows robust segmentation and counting performance. Colors in prediction masks are random and are used to represent instances. Colors may repeat. Best viewed in color and zoomed.}
\vspace{-0.6cm}
\label{results}
\end{figure*}
\subsection{Implementation Details}
\paragraph{Network Architecture.} Since we want to highlight the contribution of shape priors and boundary setting, we choose to adopt a standard fully convolutional network (FCN) introduced at \cite{ronneberger2015u}. The network consists of an encoder with eight blocks, each consists of two convolution layers followed by batch normalization and rectified linear unit (ReLU) layers. After each block we apply a $2 \times 2$ max pooling layer with a stride of 2. The encoder captures 3 channel inputs, and yields 1024 channel output. The decoder is also formed with eight blocks, each consists of feature map upsampling, two up convolution layers which halve the number of channels followed by batch normalization and ReLU layers. The output at each decoder block is concatenated to the corresponding encoder block. At the final layer, we use a $1 \times 1$ convolution layer to map 64 channel output to the number of classes.
\vspace{-0.34cm}
\paragraph{Training and Evaluation Setup.} We train our network from scratch using 90/5/5 data split with Adam optimizer \cite{kingma2014adam}, starting learning rate of 0.001, and cosine annealing scheduler \cite{loshchilov2016sgdr}. Random flips and normalization transforms are applied to the training input. We let the networks train on a single NVIDIA GTX 1080 for 25,000 iterations or until convergence, whichever comes first. For metrics, we report Mean Absolute Error (MAE) for counting and Mean Intersection over Union (mIoU) for segmentation. MAE score calculates the sum of absolute differences between count ground truth and predictions, divided by the number of examples. Count predictions are found using the number of connected components \cite{dillencourt1992general} computed on the segmentation prediction mask. mIoU computes the ratio between intersection and union between prediction and ground truth masks. We also consider the inverse MAE to mIoU ratio $Q_{cs}$ to indicate how well a model does on both metrics. $Q_{cs}$ measures the trade-off between counting and segmentation performances, and is used as a joint performance indicator. During training, models with best MAE, best mIoU, and best $Q_{cs}$ are saved. For models that incorporate segmentation, validation and testing is preformed on fully supervised images, and models are chosen using the best $Q_{cs}$. For counting specific methods, models are selected based on the best MAE results. Formal formulation of reported metrics are defined by
\begin{equation}
    \begin{aligned}
        &\text{MAE} = \frac{1}{n} \sum_{i=1}^{n} |\Tilde{c_i} - c_i| \text{ ,}\\
        &\text{mIoU} = \frac{1}{n} \sum_{i=1}^{n} \frac{y_i \cap \Tilde{y}_i}{y_i \cup \Tilde{y}_i} \text{ ,}\\
        &Q_{cs} = \frac{1}{\text{MAE}} * \text{mIoU.}
    \end{aligned}
    \label{metrics}
\end{equation}
Where $c$, $\Tilde{c}$ represent true and predicted counts in image $i$, and $n$ indicates the number of examples in the dataset.
\subsection{Baselines}
We compare our method to SOTA in counting \cite{ribera2019locating}, joint counting and segmentation \cite{laradji2018blobs}, and semantic segmentation algorithms \cite{ronneberger2015u}. All baselines were trained from scratch to ensure fair comparison. The original formulation in \cite{ribera2019locating} was unable to learn meaningful counts in our data caused by a ReLU layer at the regressor branch that zeros estimated counts. Instead, we modify \cite{ribera2019locating} (referred to as m\cite{ribera2019locating}) by using a Parametric ReLU, which learns an additional parameter to better handle negative values. It is possible that further tuning of hyperparameters is necessary for better performance. U-Net with point supervision baseline was trained with adjusted class weights to allow better learning. Also, important to note that we recognize that LC-FCN \cite{laradji2018blobs} does not aim to segment images, but since its approach is similar enough and the lack of other comparable works, we slightly modified its code to output segmentation masks and included results in both metrics. 
\subsection{Results}
Table \ref{allresults} presents  comparisons between baselines and our method for counting and segmentation metrics. As can be observed, our method outperforms \cite{ribera2019locating}, m\cite{ribera2019locating}, and LC-FCN \cite{laradji2018blobs} in those metrics. We see superior counting performance against \cite{ribera2019locating}, which proved unable to correctly count cranberries in CRAID images, although it performed significantly better on other datasets. The comparison to LC-FCN can also be seen in Figure~\ref{results}, where better separability between objects results in better counting, and more accurate shape results in better segmentation performance. Notice that the segmented blobs maintain  elliptical shapes, compared to irregular shaped blobs produced by \cite{laradji2018blobs}.
\subsection{Ablation}
We explore the contribution of each added module in our proposed method and compare them to the SOTA methods in counting and segmentation. We find that using known shape priors as a blob structuring indicator dramatically improve segmentation performance. While using $\mathcal{L}_{Convex}$ shows better results on segmentation, $\mathcal{L}_{Circ}$ provides better outcome overall with the highest inverse MAE to mIoU ratio. Table \ref{allresults} also shows that adding a count loss to segmentation and split losses boosts counting precision but degrades segmentation performance. The results also show that count loss always degrades overall results when paired with shape priors. We also examine how shape cues compare to color cues for our network. Typically, color cues are strong indicators in similar objects, which is a challenge in agriculture applications as color is a dynamic feature varying between seasons. The last row of Figure~\ref{results} shows how the network handles leaves around cranberries that redden during late fruit ripening period. It can be seen that while there are many red leaves in the scene, the majority are predicted as background by the network. 
\subsection{Conclusion}
In this paper, we present a novel approach to count and segment objects utilizing point supervision and shape priors. We propose the Triple-S network that employs our selective watershed algorithm, and shape loss functions to encourage convex and circular object masks. 
We present a first of its kind publicly available dataset and software toolkit for supporting precision agriculture in cranberry fields. The approach can be extended to other crops such as blueberries, grapes, and olives.
\vspace{-0.25cm}
{\small
\paragraph{Acknowledgements} This project was sponsored by the USDA NIFA AFRI Award Number: 2019-67022-29922. We thank David Nuhn who assisted in data collection. We acknowledge Aditi Roy at Siemens Corporate for conversations on segmentation baselines.
}

\clearpage

{\small
\bibliographystyle{ieee_fullname}
\bibliography{finding_berries.bib}
}

\end{document}